\documentclass[letterpaper]{article} % DO NOT CHANGE THIS
\usepackage{aaai25}  % DO NOT CHANGE THIS
\usepackage{times}  % DO NOT CHANGE THIS
\usepackage{helvet}  % DO NOT CHANGE THIS
\usepackage{courier}  % DO NOT CHANGE THIS
\usepackage[hyphens]{url}  % DO NOT CHANGE THIS
\usepackage{graphicx} % DO NOT CHANGE THIS
\usepackage{natbib}  % DO NOT CHANGE THIS AND DO NOT ADD ANY OPTIONS TO IT
\usepackage{caption} % DO NOT CHANGE THIS AND DO NOT ADD ANY OPTIONS TO IT
\usepackage{multirow}
\usepackage{enumitem}
\usepackage{amssymb}
\usepackage{amsfonts}
\usepackage{amsthm}
\usepackage{amsmath}

\usepackage{algorithm}
\usepackage{algorithmic}
\usepackage{cleveref}
\usepackage{multirow}
\usepackage{xcolor}
\usepackage{amsfonts}

\usepackage{newfloat}
\usepackage{listings}
\DeclareCaptionStyle{ruled}{labelfont=normalfont,labelsep=colon,strut=off} % DO NOT CHANGE THIS
\floatstyle{ruled}
\newfloat{listing}{tb}{lst}{}
\floatname{listing}{Listing}
\nocopyright
\title{Sequence Matters: Harnessing Video Models in 3D Super-Resolution}
\author{
    Hyun-kyu Ko\textsuperscript{\rm 1}\equalcontrib,
    Dongheok Park\textsuperscript{\rm 2}\equalcontrib,
    Youngin Park\textsuperscript{\rm 3},
    Byeonghyeon Lee\textsuperscript{\rm 1},
    Juhee Han\textsuperscript{\rm 1},\\
    Eunbyung Park\textsuperscript{\rm 1,2}\thanks{Corresponding author.}
}

\affiliations {
    \textsuperscript{\rm 1}Department of Electrical and Computer Engineering, Sungkyunkwan University\\
    \textsuperscript{\rm 2}Department of Artificial Intelligence, Sungkyunkwan University\\
    \textsuperscript{\rm 3}Visual Display Division, Samsung Electorics\\
}

\usepackage{bibentry}
\begin{document}

\maketitle

\begin{abstract}
3D super-resolution aims to reconstruct high-fidelity 3D models from low-resolution (LR) multi-view images. Early studies primarily focused on single-image super-resolution (SISR) models to upsample LR images into high-resolution images. However, these methods often lack view consistency because they operate independently on each image. Although various post-processing techniques have been extensively explored to mitigate these inconsistencies, they have yet to fully resolve the issues. In this paper, we perform a comprehensive study of 3D super-resolution by leveraging video super-resolution (VSR) models. By utilizing VSR models, we ensure a higher degree of spatial consistency and can reference surrounding spatial information, leading to more accurate and detailed reconstructions. Our findings reveal that VSR models can perform remarkably well even on sequences that lack precise spatial alignment. Given this observation, we propose a simple yet practical approach to align LR images without involving fine-tuning or generating `smooth' trajectory from the trained 3D models over LR images. The experimental results show that the surprisingly simple algorithms can achieve the state-of-the-art results of 3D super-resolution tasks on standard benchmark datasets, such as the NeRF-synthetic and MipNeRF-360 datasets.
\noindent
{\color{black} \rule{\linewidth}{0.2mm}}
Project Page: \url{https://ko-lani.github.io/Sequence-Matters}
\end{abstract}

\section{Introduction}

Recent advancements in 3D reconstruction from multi-view images, e.g., Neural Radiance Fields (NeRF) and 3D Gaussian Splatting (3DGS), have demonstrated outstanding performance across various tasks, such as novel view synthesis~\cite{mildenhall2021nerf, muller2022instant, chen2022tensorf, fridovich2022plenoxels, kerbl20233d} and surface reconstruction~\cite{wang2021neus, yariv2021volume, yariv2023bakedsdf, guedon2024sugar, huang20242d, fan2024trim}. In addition, these techniques have proved highly effective in creating 3D scenes and assets when combined with the generative model approaches~\cite{poole2022dreamfusion, liu2023zero}. The versatility of these methods and their ability to generate accurate and detailed 3D models have broadened their applicability to various tasks~\cite{wang2024dust3r, yu2024mip}.
 
Utilizing high-quality or high-resolution multi-view input images is crucial for obtaining high-fidelity 3D models from these techniques. However, meeting this requirement in real-world settings is often infeasible due to various constraints, e.g., equipment limitations or adverse environmental conditions. To overcome these challenges, several recent studies have investigated the 3D super-resolution task, which aims to generate high-fidelity 3D models from low-resolution multi-view images~\cite{wang2022nerf, Han2023super, yoon2023cross, lin2024fastsr, feng2024zs, lee2024disr, wu2024rafe, feng2024srgs, shen2024supergaussian, yu2024gaussiansr}. The early approaches have utilized single-image super-resolution (SISR) models. They first upscale low-resolution (LR) input images to high-resolution (HR) images and then apply NeRF or 3DGS techniques to represent the 3D models. However, they face a critical limitation; the generated HR images usually lack 3D consistency since the input view images are processed individually. Although numerous works have improved 3D consistency using refinement stages, these solutions introduced additional computational complexity and could not fully resolve the problems.

A recent work~\cite{shen2024supergaussian} has explored the use of Video Super-Resolution (VSR) models~\cite{xu2024videogigagan} to improve the 3D consistency. Inspired by the latest studies showing video generative models can achieve highly accurate 3D spatial consistency across the generated video frames~\cite{voleti2024sv3d, zuo2024videomv}, it repurposes VSR models to upsample LR multi-view images. This approach first constructs a low-resolution 3D representation using 3DGS from LR input images and then generates an LR video (a sequence of multi-view LR images) rendered from a `smooth' camera trajectory. This VSR-friendly `smooth' LR video serves as the input for the VSR model, and it is upscaled to an HR video (a sequence of multi-view HR images) from which the HR 3D model is subsequently produced.
 
While promising, the empirical evaluation has revealed certain limitations of this approach. The distribution shift between the training data (natural LR videos) and the testing data (the rendered LR videos from 3D models, e.g., 3DGS) negatively impacted the pre-trained VSR models. The rendered images from 3DGS frequently introduce stripy or blob-like artifacts, degrading the VSR models' performance. Although fine-tuning the VSR models on the rendered images from 3DGS could mitigate the distribution mismatch issue, posed multi-view image data is not abundant compared to natural videos, which limits the generalization performance. In addition, it is time-consuming and computationally heavy since it requires training 3DGS to obtain 3D representations for rendering input images. Consequently, the up-to-date 3D super-resolution techniques utilizing the VSR models have yet to demonstrate superior results over those leveraging SISR models~\cite{lim2017enhanced, wang2018esrgan, liang2021swinir}.
 
In this work, we propose a method that ensures the VSR models receive their desired input without fine-tuning them. We have made two critical observations regarding VSR models: 1. The artifacts introduced by the rendered images substantially comprise the performance, and 2. The VSR models maintain strong performance even when input videos do not adhere to `smooth' camera trajectories.  
Given these critical observations, we propose surprisingly simple yet effective algorithms to order training datasets into structured 'video-like' sequences. These 'video-like' sequences lead to improved VSR results, while eliminating the need for fine-tuning VSR models as they are composed of ground truth LR images, ensuring freedom from stripy or blob-like artifacts.
The experimental results have shown that our proposed algorithms achieved state-of-the-art results on the NeRF synthetic and Mip-NeRF 360 datasets, underscoring their efficacy and robustness. Our key contributions are summarized below.

\begin{itemize}
\item We propose a novel method that leverages VSR models to bridge the gap between low-resolution and high-resolution images. By generating input video sequences that are sufficiently ‘smooth’ and exhibit minimal artifacts, we optimize their suitability for VSR models.
\item We propose surprisingly simple yet effective ordering algorithms, demonstrating superior performance compared to the existing prior arts.
%\item We demonstrate that VSR models can effectively handle ‘non-smooth’ sequences, showing superior performance compared to SISR models by efficiently leveraging neighboring frames.
\item Our method achieves state-of-the-art performance on both object-level and scene-level datasets, including the NeRF Synthetic and Mip-NeRF 360 datasets, highlighting the robustness and effectiveness of our approach in both object and scene datasets.
\end{itemize}

\section{Related Work}

\subsubsection{Novel View Synthesis}
Novel view synthesis (NVS) is the task of synthesizing images from novel viewpoints given multi-view images. With the rise of deep learning, Neural Radiance Fields (NeRF)~\cite{mildenhall2021nerf} achieved remarkable results by learning a continuous function of the scene with MLP and can render the novel views with a volumetric renderer. In contrast to NeRF and its variants~\cite{mildenhall2021nerf, barron2021mip, barron2022mip, muller2022instant, chen2022tensorf, fridovich2022plenoxels}, which learns the implicit 3D representation of the scene, 3D Gaussian Splatting (3DGS)~\cite{kerbl20233d} learns the point cloud-based explicit 3D representation. Since 3DGS employs explicit representation and renders images through rasterization, it achieves real-time rendering without compromising the quality of rendered images. However, to learn high-fidelity 3D representation, these neural fields require high-resolution images, which is not always guaranteed in real-world environments.
In this work, we study 3D super-resolution task, where we build 3D representations given the only LR images.

\subsubsection{3D Super-resolution}
\label{related:3DSR}
Despite the great success of 3D neural fields in various applications, it is challenging to reconstruct high-resolution (HR) radiance fields using low-resolution datasets. Recently, several studies~\cite{wang2022nerf, feng2024zs} have attempted to achieve 3D super-resolution using super-sampling techniques without the guidance of off-the-shelf models, such as image restoration or generative models. In contrast, another line of research~\cite{Han2023super, yoon2023cross, lin2024fastsr, lee2024disr, feng2024srgs, yu2024gaussiansr, xie2024supergs} has focused on improving the resolution of 3D representations with the aid of these established models. They utilize SISR models to upsample low-resolution images and incorporate additional modules or techniques to enhance multi-view consistency across the upsampled images. Recent works in this line~\cite{Han2023super, yoon2023cross, feng2024srgs} utilize SISR models to upsample training datasets. On the other hand, another work~\cite{lin2024fastsr} upsamples rendered images for fast inference speed. Additionally, other studies~\cite{lee2024disr, yu2024gaussiansr} employ a latent diffusion model (LDM)~\cite{rombach2022high} and score distillation sampling (SDS) loss~\cite{poole2022dreamfusion} to achieve 3D super-resolution.

A recent work~\cite{shen2024supergaussian}, closely related to ours, leverages a VSR model as an upsampler for the LR dataset. Unlike SISR models, which often do not consider other frames during super-resolution, VSR models reference adjacent frames, thereby enhancing multi-view consistency. Specifically, this work starts by training the 3DGS with LR images and renders LR video frames from the trained LR 3DGS. Subsequently, with the LR video rendered from the LR 3DGS, the VSR model generates the training dataset of HR 3DGS. However, the distribution of the LR dataset differs from that of the rendered LR video, thereby introducing stripy or blob-like artifacts in the upsampled video. Fine-tuning the VSR model with rendered LR videos can mitigate this distribution shift, but it is a time-consuming process, as it involves extensive training and rendering of LR 3DGS to generate the fine-tuning dataset (LR rendered videos). In this work, we propose a method that does not involve additional finetuning or training 3DGS on LR images to render `smooth' video.

\subsubsection{Video Super-resolution}
Video super-resolution (VSR) has evolved from advancements in image super-resolution~\cite{wang2018esrgan, liang2021swinir, zhang2021designing, chen2023activating, tian2024image}, generating high-resolution video frames by utilizing information from adjacent frames to enhance the current frame's resolution.
BasicVSR~\cite{chan2021basicvsr} introduced a bidirectional propagation approach to achieve balanced references from both directions and compute optical flow from features rather than images for more accurate alignment. BasicVSR++~\cite{chan2022basicvsr++} built on this by using second-order grid propagation, which extracts features through multiple stages and incorporates information from non-adjacent frames, enhancing robustness to occlusion.
VRT~\cite{liang2024vrt} advanced VSR by combining recurrent model-based approaches with transformer structures and PSRT~\cite{shi2022rethinking} proposed patch alignment, which aligns image patches rather than individual pixels, utilizing self-attention to enhance alignment and performance.
Additionally, IART~\cite{xu2024enhancing} introduced a neural network-based resampling strategy, employing sinusoidal positional encoding and a transformer-based coordinate network to preserve high-frequency details and reduce spatial distortions. In this work, we harness the recent VSR models' capability to improve the multi-view 3D super-resolution tasks.

\begin{figure}[t]
\centering
\includegraphics[width=0.95\columnwidth]{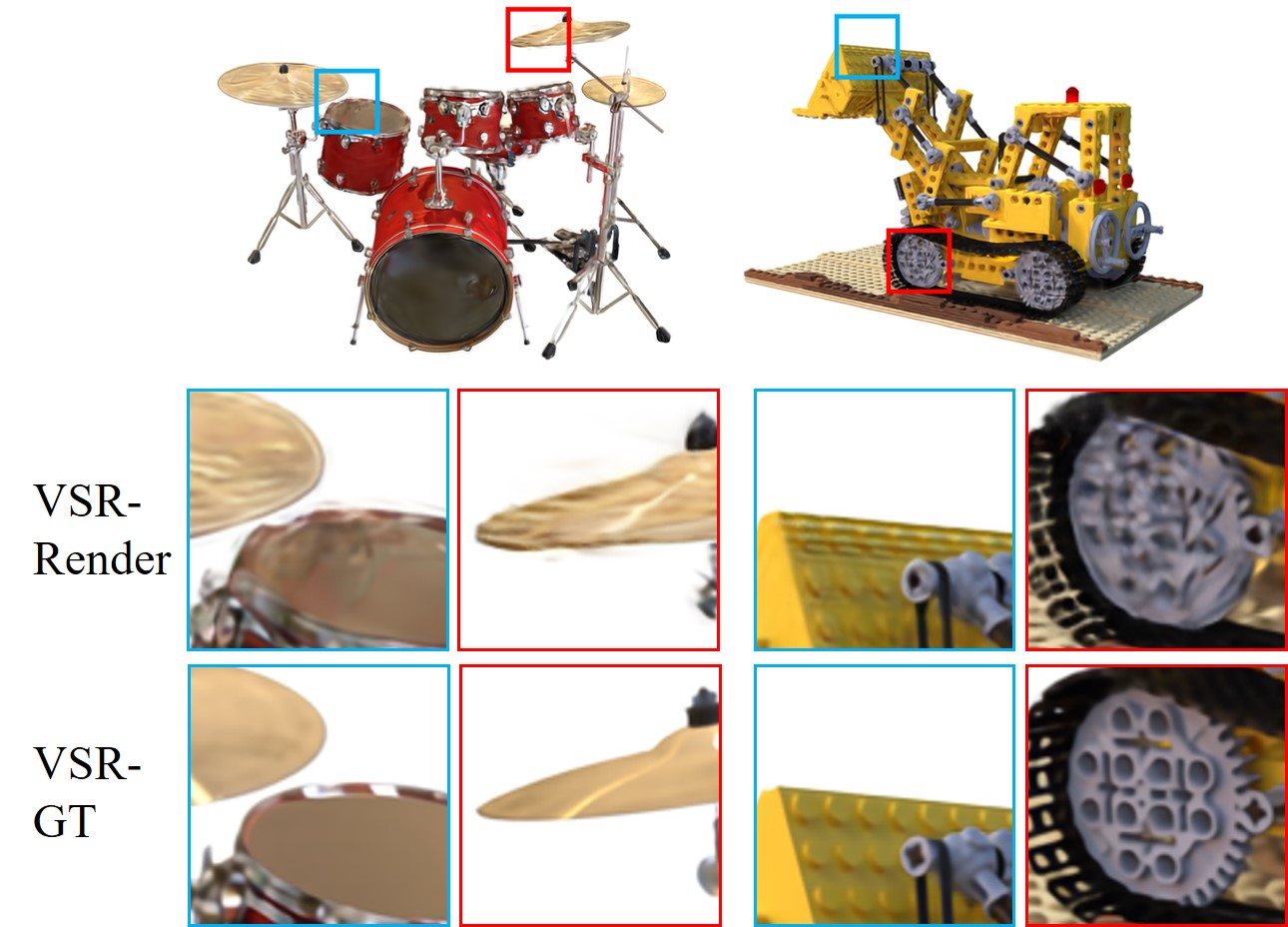}
% \caption{Illustration of stripy or blob-like artifact generated in VSR outputs of LR video rendered from 3DGS. `VSR-Render' refers to the VSR outputs of rendered images, and `VSR-GT' refers to the VSR outputs of low-resolution GT images (Testing dataset of Blender dataset).}
\caption{Illustration of stripy or blob-like artifacts generated in VSR outputs of LR videos rendered from 3DGS. `VSR-Render' shows the VSR outputs of the LR rendered videos, while `VSR-GT' displays the VSR outputs of the ground truth (GT) LR videos.}
\label{fig:stripy}
\end{figure}

\section{Method}

\subsection{Rendering Artifacts}
\label{sub:rendering_video_in_novel_view}
In most multi-view datasets and image acquisition scenarios, images are hardly spatially ordered (except for a few cases, e.g., a monocular camera captures a sequence of images and records the time they are taken), which are unfavored for VSR models.
A straightforward approach to obtaining the spatially ordered images from LR multi-view images would involve obtaining 3D representations with LR images, such as 3DGS, and rendering a smooth video from them \cite{shen2024supergaussian}.
However, this approach introduces a significant problem due to the mismatch between the rendered images from the 3DGS trained on LR multi-view images and images from LR video datasets on which the VSR model was trained, e.g., bicubic downsampled from HR videos. This mismatch often results in blob-like or stripy artifacts from 3DGS, which degrades the performance of VSR models. 
\citet{shen2024supergaussian} partially addressed this issue by finetuning the VSR models on the images rendered from 3DGS. While effective, it demands training 3DGS for each training instance, which increases significant computational complexity. Furthermore, the multi-view image dataset is less abundant than natural videos, limiting the generalization performance of finetuned VSR models.

We have investigated these blob-like, stripy artifacts of the rendered images from the 3DGS models trained on LR multi-view images. These are primarily observed in the regions where high-fidelity information from HR images is lost in the LR images. The VSR models take the damaged images as inputs and upsample them, preserving or often magnifying the artifacts, which significantly degrades the output quality. As shown in Fig.~\ref{fig:stripy}, the regions with lost details in the LR images become severe artifacts in the upsampled images.

\subsection{A Simple Greedy Algorithm}
\label{sub:aligning_images_with_pose_and_feature}

In Sec.~\ref{sub:rendering_video_in_novel_view}, we demonstrated the limitations of training 3DGS with low-resolution (LR) images to obtain a `smooth' video. This section explores alternative approaches that exploit the raw unordered LR multi-view images to create a video-like sequence.

Determining the most desirable order for generating video-like sequences from unordered LR datasets is challenging due to the absence of clear criteria, such as `how video-like' a sequence should be or what makes a sequence a `good video' for VSR models. 
Our objective is to arrange a sequence of images to maximize the quality of the high-resolution (HR) images produced by VSR models. However, the absence of ground-truth HR images, as defined by the problem, makes it infeasible to establish a clear objective function.
We consider a rather simple approach: a `good' video is a sequence in which each frame is `similar' to its adjacent frames, ensuring a smooth visual flow.

\begin{algorithm}[t]
    \caption{A Simple Greedy Algorithm}
    \label{alg:a_simple_greedy_algorithm}
    \textbf{Input}: A set of unordered images, $I = \{ I_j \}_{j=1}^N$ \\
    \textbf{Output}: An ordered sequence of images, $S$
    
    \begin{algorithmic}[1] %[1] enables line numbers
        \STATE $S_1 = I_1, \quad $ $I \leftarrow I \setminus {\{I_1\}}$
        \FOR {$j \gets 1$ to $N-1$}
            \STATE $S_{j+1} = \underset{I_k \in I}{\operatorname{argmin}} \operatorname{sim}(S_j,I_k)$
            \STATE $I \leftarrow I \setminus \{S_{j+1}\}$
        \ENDFOR
        \RETURN $S$
    \end{algorithmic}
\end{algorithm}

Although these criteria were well-defined, finding the optimal sequence remains NP-hard due to the combinatorial nature of the problem. Our investigation, however, demonstrated that VSR models are sufficiently robust to non-optimal sequences, effectively utilizing distant multi-view references for upsampling. Given the observation, we propose a simple yet practical greedy algorithm (Alg.~\ref{alg:a_simple_greedy_algorithm}). Starting from an initial image $S_1$, it repeatedly finds the next image by using the nearest neighbor based on the similarity score $\operatorname{sim}(\cdot,\cdot)$.

We explore two similarity measures, camera poses and visual features. By utilizing camera poses, we can spatially connect images that are close to each other to form a video.
Although this approach is conceptually sound, it may lack generalizability across diverse datasets, such as Mip-NeRF 360 dataset, which is not object-centric (i.e., the images are not all focused on the same object). As an alternative, we explore the visual feature-based similarity. We evaluated multiple feature extractors~\cite{lowe2004distinctive}~\cite{rosten2006machine}~\cite{bay2006surf}~\cite{calonder2010brief} and found that ORB (Oriented FAST and Rotated BRIEF) feature~\cite{rublee2011orb} offers a balance of computational efficiency and robustness in feature matching.

\begin{figure*}[ht]
\centering
\includegraphics[width=0.95\textwidth]{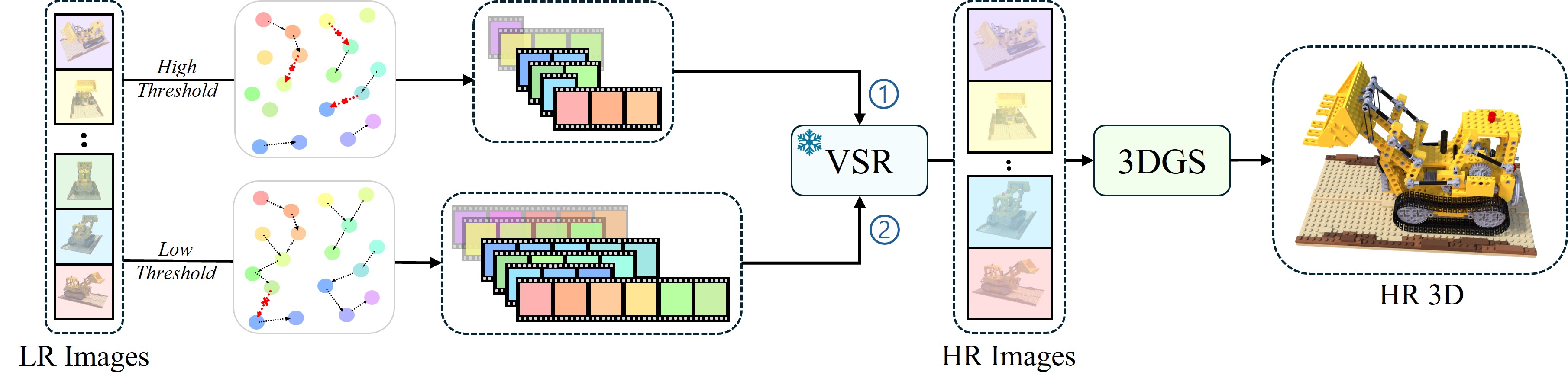} 
\caption{Overview of the proposed method. Given LR multi-view images, we generate subsequences (Sec.~\ref{sub:adaptive_subsequence_generation}) starting from each image using a simple greedy algorithm (Sec.~\ref{sub:aligning_images_with_pose_and_feature}) and these subsequences are bounded by multiple thresholds (Sec.~\ref{sub:multi_threshold_subsequence}). Finally, we train a 3DGS model for 3D reconstruction using the upsampled HR images.}
\label{fig:fig2}
\end{figure*}

\subsection{Adaptive-Length Subsequence}
\label{sub:adaptive_subsequence_generation}

\begin{figure}[t]
\centering
\includegraphics[width=0.9\columnwidth]{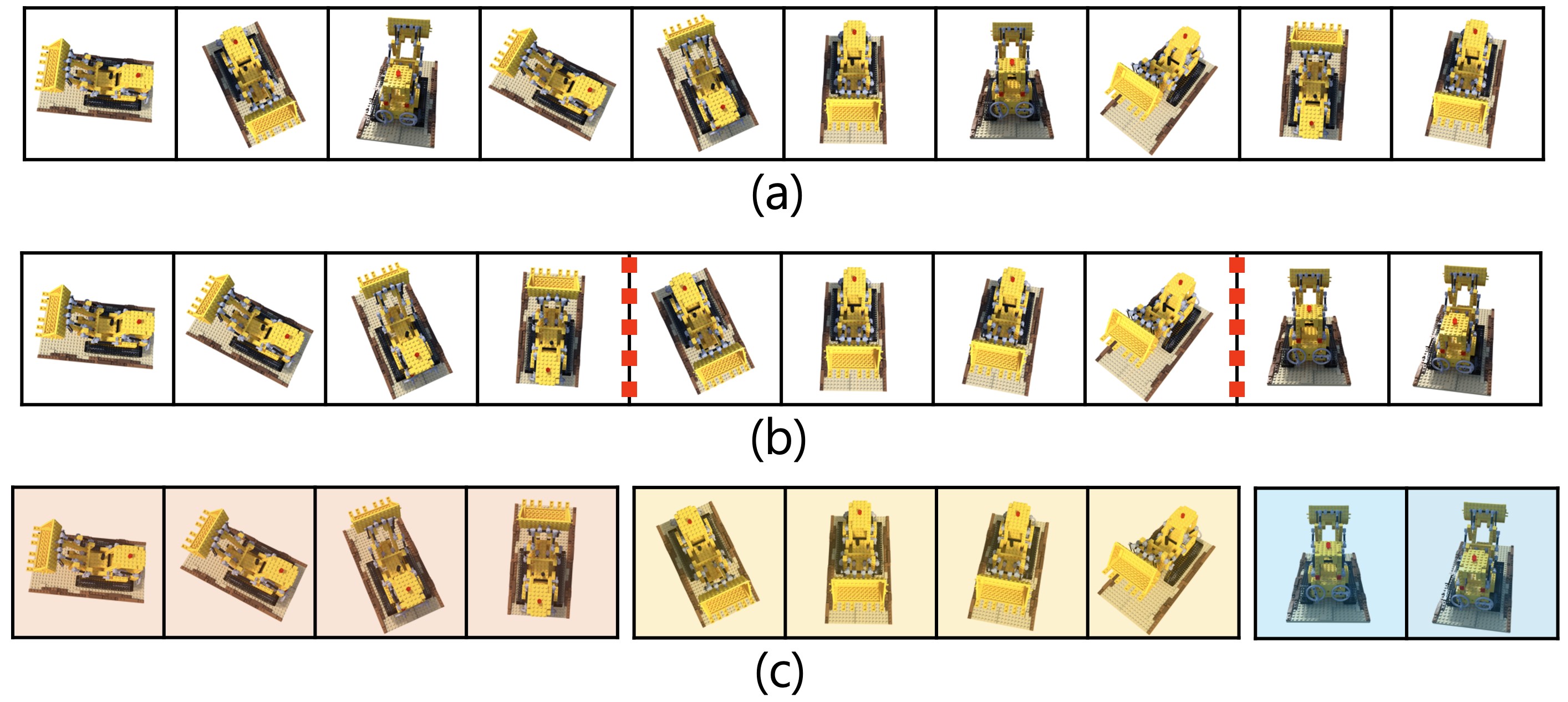}
\caption{Illustration of subsequence generation. (a) is an unordered multi-view image dataset. (b) is the result of using a simple greedy algorithm, Alg.~\ref{alg:a_simple_greedy_algorithm}. (c) highlights misalignments incurred by the algorithm, and we propose to split it into subsequences based on a pose difference threshold (red dotted line) between consecutive frames.}
\label{fig:subsequence}
\end{figure}

While promising, the proposed simple greedy algorithm faces two challenges when connecting all images into a single video sequence. First, the resulting sequence often exhibits abrupt transition due to the inherent weaknesses of greedy algorithms (illustrated in Fig.~\ref{fig:subsequence}-(b)). For instance, the nearest neighbor of $S_{k}$ may have already been included in the processed list ($S_1, ..., S_{k-1}$), forcing the selection of a far-distant image.

Second, the results are highly influenced by the choice of the initial image $S_1$. To address these challenges, we improve the algorithm by 1. stopping building sequence when the similarity score does not meet a certain threshold and 2. creating multiple subsequences starting from each image in the dataset. 

\begin{algorithm}[t]
    \caption{Adaptive-length Subsequening}
    \label{alg:adaptive_length_subsequence}
    \textbf{Input}: A set of unordered images, $I = \{ I_j \}_{j=1}^N$ \\
    \textbf{Output}: Multiple ordered sequences, $\{S^{(j)}\}_{j=1}^N$
    
    \begin{algorithmic}[1] %[1] enables line numbers
    \FOR {$i \gets 1$ to $N$}
        \STATE $S_1^{(i)} = I_i, \quad $ $I \leftarrow \{ I_j \}_{j=1}^N \setminus {\{I_i\}}$
        \FOR {$j \gets 1$ to $N-1$}
            \STATE $S_{j+1}^{(i)} = \underset{I_k \in I}{\operatorname{argmin}} \operatorname{sim}(S_j^{(i)},I_k)$
            \IF {$\operatorname{sim}(S_j^{(i)},S_{j+1}^{(i)}) < \epsilon$}
            \STATE $S^{(i)} = S^{(i)}_{1:j}$ \hfill {$//$ The length of $S^{(i)}$ becomes $j$}
            \STATE \textbf{break}
            \ENDIF
            \STATE $I \leftarrow I \setminus \{S_{j+1}^{(i)}\}$
        \ENDFOR
    \ENDFOR
    \RETURN $\{S^{(j)}\}_{j=1}^N$
    \end{algorithmic}
\end{algorithm}

Alg.~\ref{alg:adaptive_length_subsequence} describes the detailed algorithm, and each subsequence $S^{(j)}$ is an ordered sequence starting from the initial image $I_j$, and each subsequence has different lengths. Finally, we apply VSR models to upsample the subsequence and aggregate the outputs to generate the final upsampled sequence $\hat{I} = \{ \hat{I}_j \}_{j=1}^N$ as follows,
\begin{equation}
    \hat{S}^{(j)} = \operatorname{VSR}(S^{(j)}), \quad \hat{I} = \operatorname{agg}(\{\hat{S}^{(j)}\}_{j=1}^N),
\end{equation}
where $\operatorname{agg}$ is an aggregate operator that takes multiple upsampled sequences as input and produces the final upsampled sequence. During aggregation, it removes redundant images, retaining only the image from the earliest subsequence ($|I|=|\hat{I}|=N$).

Each similarity measure has its own limitations. Pose similarity suffers from the different orientations of cameras. The proximity in camera position does not account for the fact that the cameras may be facing in different directions, leading to connections between unrelated images. On the other hand, feature similarity can lead to incorrect alignments (significantly different image pairs often have a high similarity score).
We observed that when dividing sequences into subsequences, pose and feature can complement each other. For example, we can use feature similarity for $\operatorname{sim}$ in line 4 and pose similarity for $\operatorname{sim}$ in line 5 of Alg.~\ref{alg:adaptive_length_subsequence}.

\subsubsection{Multi-threshold Subsequence Generation}
\label{sub:multi_threshold_subsequence}

We generate subsequences from each multi-view image in the dataset based on a uniform threshold.

However, applying a uniform threshold across all sequences can lead to inefficiencies. Setting a high threshold imposes strict constraints, ensuring that only very closely related images are connected, which results in smoother sequences but shorter sequences. On the other hand, a lower threshold ensures longer sequences, which often compromises the smoothness of the resulting sequences.

To leverage both advantages, we introduce a multi-threshold generation approach as illustrated in Fig.~\ref{fig:fig2}. Initially, we apply a high threshold for creating the subsequences, prioritizing the smoothest subsequences. These subsequences are then processed through the VSR model for upsampling. However, since not all images can be processed by a high-threshold approach (note that VSR models require a certain number of frames), we then lower the threshold in the next iteration, creating more relaxed and less smooth subsequences to include the remaining images. Please refer to more detailed algorithms in the Appendix.

\subsection{Training Objective}
\label{sub:subpixel_constraints}
We use a VSR model to upsample LR images to enhance multi-view consistency. However, generated high-frequency details are not always consistent across different views, which leads to degrade the quality of 3D reconstruction. Following~\cite{wang2022nerf, feng2024srgs}, we use sub-pixel constraints to regularize inconsistent high-frequency details.
%Unlike NeRF-SR~\cite{wang2022nerf} and SRGS~\cite{feng2024srgs}, which leverage average kernel to downsample HR images, we downsample HR images through bicubic interpolation. 
In practice, since bicubic interpolation is used to generate the low-resolution (LR) dataset, we also utilize bicubic interpolation when downsampling the sub-pixels.
The sub-pixel loss $\mathcal{L}_{\text{sp}}$ is LR 3DGS loss calculated between LR images and downsampled rendered images. Then, the final loss of our framework is expressed as below:
\begin{equation}
\label{final_loss}
    \mathcal{L} = \lambda_{\text{ren}} \mathcal{L}_{\text{ren}} + (1-\lambda_{\text{ren}}) \mathcal{L}_{\text{sp}},
\end{equation}
% where $\lambda^{3dgs}$ HR 3DGS loss. Please see Appendix. \textbf{F} for more details of the loss.
where $\mathcal{L}_{\text{ren}}$ HR 3DGS loss. Please see Appendix. \textbf{F} for more details.

\section{Experiment}

\subsection{Setup}
\subsubsection{Datasets}
We use the NeRF Synthetic Blender dataset~\cite{mildenhall2021nerf} and the Mip-NeRF 360 dataset~\cite{barron2022mip}. The Blender dataset consists of 8 synthetic object-centric scenes, with each scene with a resolution of 800 $\times$ 800. For our experiments, we downsampled the images with bicubic interpolation by a factor of 4 to create a low-resolution (LR) dataset (200 $\times$ 200). The Mip-NeRF 360 dataset contains 9 real-world scenes. The resolutions vary across the scenes, but each scene has a higher resolution compared to the Blender dataset. We downsampled the dataset with bicubic interpolation by a factor of 8 to create the LR dataset.

\subsubsection{Metrics}
\label{sub:metrics}
Following the previous works, we evaluate the quantitative results using PSNR, SSIM, and LPIPS. Some previous works~\cite{Han2023super, lee2024disr, wu2024rafe, shen2024supergaussian} emphasize the importance of perceptual metrics such as NIQE and LPIPS rather than fidelity metrics like PSNR. However, we prioritize the PSNR metric, as we regard the super-resolution task as a subset of reconstruction tasks, where accurate reconstruction of the original image is crucial.

\subsubsection{Background Impact on Metrics}
When measuring metrics on the Blender dataset, we follow DiSR-NeRF and RaFE by using a black background where the alpha channel value is 0, unlike NeRF-SR, which used a white background.
We observed that compositing with a white background introduces black artifacts around the edges of the images, making it difficult to obtain accurate measurements.
In our experiments, the artifacts from compositing with a white background significantly degraded the output quality (empirically, by about 0.3 to 0.4 on PSNR).
Since most of the previous works have not released their code and do not mention the background issue, we are unable to determine which background they used for their metrics.

\subsubsection{Baseline Models}
As a baseline, we examined NeRF-SR~\cite{wang2022nerf}, ZS-SRT~\cite{feng2024zs}, CROP~\cite{yoon2023cross}, FastSR-NeRF~\cite{lin2024fastsr}, DiSR-NeRF~\cite{lee2024disr}, SRGS~\cite{feng2024srgs}, GaussianSR, and SuperGaussian, following the metrics used by these models.
Unfortunately, only two models, NeRF-SR and DiSR-NeRF, have provided their codes publicly.
We have added three additional baseline methods: Bicubic, SwinIR, and Render-SR.
For Bicubic and SwinIR, we upsampled LR images using bicubic interpolation and the SwinIR model, respectively. For Render-SR, we trained 3DGS with LR images in a SuperGaussian manner and upsampled the rendered smooth video using PSRT~\cite{shi2022rethinking}.
After the upsampling process, we used 3DGS for the 3D reconstruction of these models.
For NeRF-SR, DiSR-NeRF, and the three additional baselines, rendering is conducted with a white background by default.
Note that our model and SuperGaussian are based on video super-resolution models, whereas GaussianSR, NeRF-SR, ZS-SRT, CROP, and SRGS are all based on single-image super-resolution models.
Additionally, NeRF-SR, ZS-SRT, FastSR-NeRF, CROP, and DiSR-NeRF are NeRF-based models, while SRGS, GaussianSR, SuperGaussian, and our model are based on 3DGS.

\subsubsection{Implementation Details}
We implement our method using the open-source 3D Gaussian Splatting code base.
Following the 3DGS protocol, we train both coarse and fine 3DGS models for 30,000 iterations.
To create the low-resolution (LR) dataset, we downsample the high-resolution (HR) dataset using bicubic interpolation with a downscale factor of 4.
As a VSR backbone of our model, we employed PSRT~\cite{shi2022rethinking}.
Please refer to the Appendix for further details.

\begin{figure}[ht!]
\centering
\includegraphics[width=0.95\columnwidth]{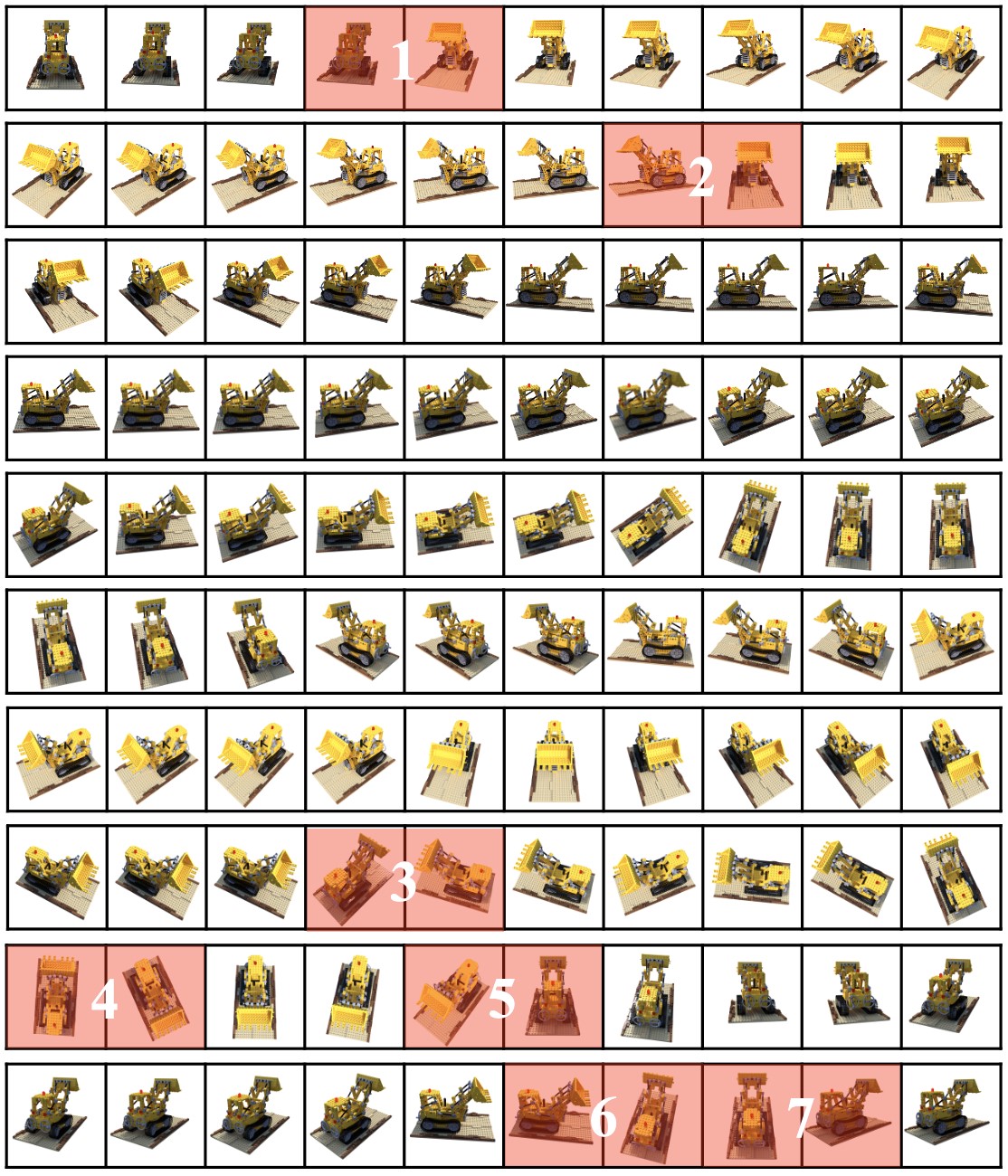} % Reduce the figure size so that it is slightly narrower than the column.
\caption{An example result from the simple greedy algorithm applied to the NeRF-synthetic dataset (Lego). Two neighboring images highlighted in red demonstrate abrupt transitions caused by misalignments.}
\label{fig:enhancing_misalignment}
\end{figure}

\subsection{Results}
\subsubsection{The Effect of The Proposed Algorithms}

Fig.~\ref{fig:enhancing_misalignment} shows a resulting sequence of the proposed simple greedy algorithm (Lego, training images from the NeRF-synthetic dataset). We used the visual feature for similarity measure in this example, and two neighboring images with red color highlight the abrupt transition between two subsequent images due to the misalignment of the algorithm. We upsampled the ordered sequence using the VSR model and calculated the PSNR of the upsampled images highlighted in Fig.~\ref{fig:enhancing_misalignment} with the ground-truth HR images. Tab.~\ref{tab:nearest_orb_comparison} shows the comparison between the simple greedy algorithm (S) and the adaptive-length subsequence algorithm (ALS).
Since ASL offers smoother transitions and is more VSR-friendly, it allows the VSR model to reference more information from neighboring images. This significantly enhances the quality of the upsampled images, demonstrating the effectiveness of the approach. 
The Tab.~\ref{tab:nearest_orb_comparison} further shows the consistent improvement of ALS over the simple greedy algorithm. First we ran the simple greedy algorithm to order the sequence and find the image pairs that their angles are more than 45 degrees. For those non-smooth image pairs, we compared the performance of the proposed ordering algorithms.

\begin{table}[ht]
\centering
\caption{The quantitative results of the proposed ordering algorithms. S: the simple greedy algorithm, ALS: the adaptive-length subsequence. L and R denote the PSNR of the left and right image in two image pairs from Fig.~\ref{fig:enhancing_misalignment}.}
\label{tab:nearest_orb_comparison}
\begin{tabular}{|c|cc|cc|}
\hline
\multirow{2}{*}{Index} & \multicolumn{2}{c|}{L}    & \multicolumn{2}{c|}{R}    \\ \cline{2-5} 
      & \multicolumn{1}{c|}{S} & ALS & \multicolumn{1}{c|}{S} & ALS \\ \hline
    1 & \multicolumn{1}{c|}{34.06} & 37.18 & \multicolumn{1}{c|}{34.53} & 35.52 \\ 
    2 & \multicolumn{1}{c|}{33.12} & 34.33 & \multicolumn{1}{c|}{34.67} & 36.63 \\ 
    3 & \multicolumn{1}{c|}{32.90} & 33.37 & \multicolumn{1}{c|}{31.62} & 34.26 \\ 
    4 & \multicolumn{1}{c|}{33.47} & 34.41 & \multicolumn{1}{c|}{33.68} & 34.29 \\ 
    5 & \multicolumn{1}{c|}{34.07} & 35.68 & \multicolumn{1}{c|}{32.77} & 35.31 \\ 
    6 & \multicolumn{1}{c|}{32.65} & 34.41 & \multicolumn{1}{c|}{32.05} & 32.77 \\ 
    7 & \multicolumn{1}{c|}{32.71} & 33.43 & \multicolumn{1}{c|}{32.68} & 34.66 \\ \hline
\end{tabular}
\end{table}

\begin{table}[ht]
    \centering
    % \fontsize{9}{11}\selectfont
    \caption{The comparison of the proposed ordering algorithms in the NeRF-synthetic dataset.}
    \label{tab:nearest_orb_comparison}
    \begin{tabular}{|l||c|c|}
        \hline
            & S & ALS\\
        \hline
        Chair & 32.11 & 32.74 \\
        Drums  & 29.74 & 30.26 \\
        Ficus & 35.31 & 35.96 \\
        Hotdog & 37.85 & 38.32 \\
        Lego  & 33.30 & 34.73 \\
        Materials & 35.24 & 35.85\\
        Mic  & 31.38 & 31.62 \\
        Ship  & 30.03 & 30.48 \\
        \hline
    \end{tabular}
\end{table}

\begin{figure*}[ht]
\centering
\includegraphics[width=\textwidth]{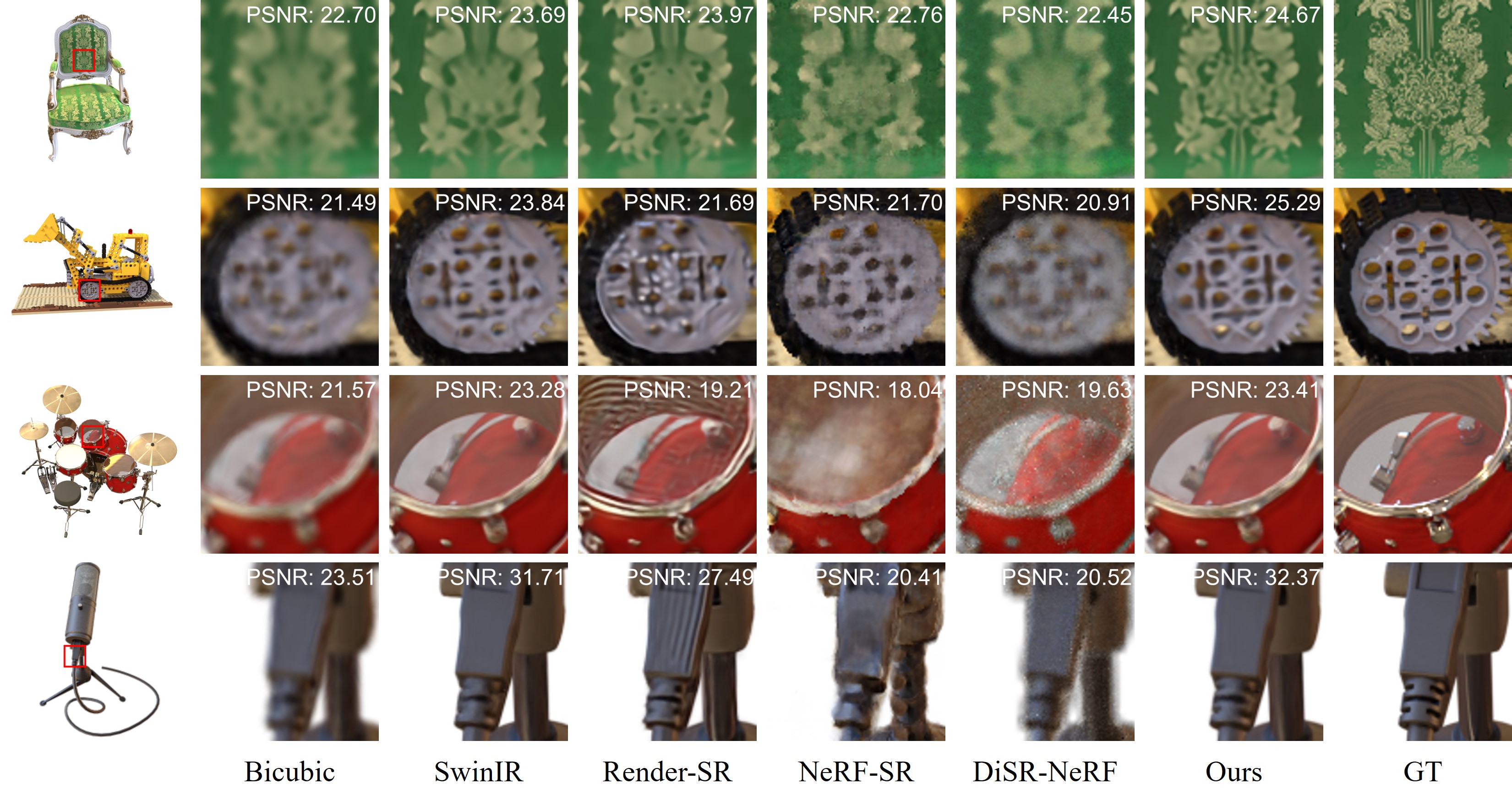} % Reduce the figure size so that it is slightly narrower than the column.

\caption{Qualitative results on the NeRF-synthetic dataset. The PSNR values against GT are embedded in each image patch. Ours have shown superior results than the existing baselines, especially for high-frequency details.}
\label{fig:qualitative_blender2}
\end{figure*}

\begin{figure}[h]
\centering
\includegraphics[width=0.95\columnwidth]{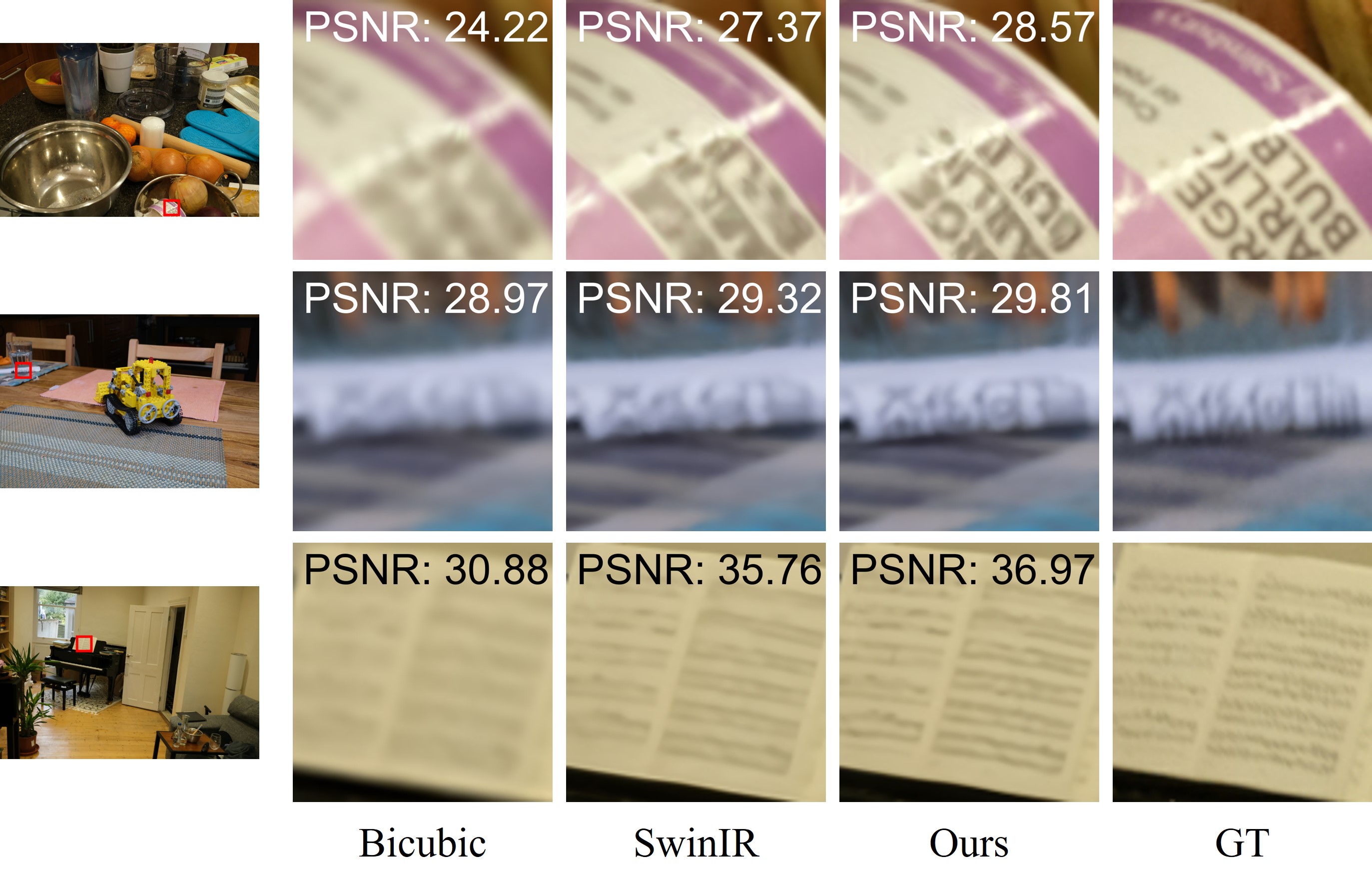} % Reduce the figure size so that it is slightly narrower than the column.
\caption{Qualitative results on Mip-NeRF 360 dataset. The PSNR values against GT are embedded in each image patch.
Ours have shown superior results than the existing baselines, especially for high-frequency details.}
\label{fig:qualitative_mip360}
\end{figure}

\begin{table}[ht]
    \centering
    \caption{Comparison of different methods for 3D super-resolution ($\times4 \rightarrow \times1$) in Blender Dataset. The numbers marked with $^{\dagger}$ are sourced from their respective paper, as the code is not available at this time.}
    \label{tab:baseline}
    % \fontsize{9}{11}\selectfont
    \begin{tabular}{|l|c|c|c|}
        \hline
        & PSNR$\uparrow$ & SSIM$\uparrow$ & LPIPS$\downarrow$\\
        \hline
        % Bicubic    & 28.64 & 0.9168 & 0.1054\\
        % SwinIR    & 31.10 & 0.9495 & 0.0564\\
        Bicubic    & 27.56 & 0.9150 & 0.1040\\
        SwinIR    & 30.77 & 0.9501 & 0.0550\\
        Render-SR & 28.90 & 0.9346 & 0.0683 \\
        \hline
        NeRF-SR      & 28.46 & 0.9210 & 0.0760\\
        ZS-SRT$^{\dagger}$      & 29.69 & 0.9290 & 0.0690\\
        \hline
        CROP$^{\dagger}$        & 30.71 & 0.9459 & 0.0671\\
        FastSR-NeRF$^{\dagger}$ & 30.47 & 0.9440 & 0.0750\\
        % DiSR-NeRF     & 25.76  & 0.8924 & 0.1212 \\
        DiSR-NeRF     & 26.00  & 0.8898 & 0.1226 \\
        SRGS$^{\dagger}$   & 30.83 & 0.9480 & 0.0560\\
        GaussianSR$^{\dagger}$    & 28.37 & 0.9240  & 0.0870 \\
        \hline
        SuperGaussian$^{\dagger}$ & 28.44 & 0.9459  & 0.0670 \\
        %Ours-ALS (wb) & 31.04 & 0.9522 & 0.0532 \\
        %Ours-S & \textbf{0} & \textbf{0} & \textbf{0} \\
        Ours-ALS & \textbf{31.41} & \textbf{0.9520} & \textbf{0.0540} \\
        \hline
        3DGS-HR  & 33.31 & 0.9695 & 0.0303\\
        \hline
    \end{tabular}
\end{table}

\subsubsection{3D Super-Resolution Results}
We provide the 3D super-resolution results, where we measure the metrics on test-view images rendered from the trained 3DGS models.
The quantitative comparison with baseline models on Nerf-synthetic in Tab.~\ref{tab:baseline}. 
Due to the space constraints, we provided the quantitative results on the Mip-NeRF 360 dataset in the Appendix.
The 3DGS-HR is the result of 3DGS trained on ground-truth HR images, which is considered as the upper bound.
The values with $\dagger$ were taken from the original papers as their codes are not publicly available.
In all baseline comparisons, the best performance is highlighted in bold.

Our methods (Ours-ALS) consistently outperformed other baseline models across all metrics. The Comparison against Render-SR and SuperGaussian, clearly highlights that the proposed methods do not suffer from stripy or blob-like artifacts.

\subsubsection{Qualitative Results}
While we showed improvements in various aspects, super-resolution tasks are notoriously challenging to improve quantitative metrics, such as PSNR. This is due to the most improvement comes from small parts of the images or high-frequency details, which conventional metrics do not accurately capture.
We provide a few qualitative results to demonstrate the effectiveness of the proposed algorithms (Fig.~\ref{fig:qualitative_blender2} and Fig.~\ref{fig:qualitative_mip360}).
We compared ours to the baseline models whose codes are available, such as NeRF-SR and DiSR-NeRF.
On the NeRF-synthetic dataset, we compared our model against Bicubic, SwinIR, Render-SR, NeRF-SR, and DiSR-NeRF.
For the Mip-NeRF 360 dataset, we compared ours with Bicubic and SwinIR, as the NeRF models for NeRF-SR and DiSR-NeRF do not perform well on the Mip-NeRF 360 dataset.
In both datasets, our model retains more high-frequency details and best reconstructs the ground truth.

\section{Conclusion}
In this paper, we introduce simple yet practical algorithms to leverage the existing VSR models to improve the 3D super-resolution task.
We proposed a simple greedy algorithm to efficiently generate a desirable sequence for the VSR models.
We further improved the resulting sequence with the adaptive-length sequence technique.
Using the proposed algorithms, we addressed the issue of stripy or blob-like artifacts caused by the trained 3D models on LR images and achieved promising performance without involving fine-tuning the VSR models.
The experimental results demonstrated the effectiveness of the proposed algorithms, showing the state-of-the-art results on standard benchmark datasets. We believe this work paves the way for more robust and efficient 3D super-resolution techniques by rethinking how to leverage VSR models, offering valuable insights for future research and development in this field.

\section*{Acknowledgements}
This work was supported by Institute of Information \& communications Technology Planning \& Evaluation(IITP) grant funded by the Korea government(MSIT) (RS-2019-II190421, Artificial Intelligence Graduate School Program(Sungkyunkwan University)) and the National Research Foundation (NRF) grant (RS-2024-00337548). This work was also supported by the Culture, Sports, and Tourism R\&D Program through the Korea Creative Content Agency grant funded by the Ministry of Culture, Sports and Tourism in 2024 (Project Name: Research on neural watermark technology for copyright protection of generative AI 3D content, RS-2024-00348469), and Samsung Research Funding \& Incubation Center of Samsung Electronics under Project Number SRFC-IT2401-01.

\bibliography{aaai25}

\begin{thebibliography}{48}
\providecommand{\natexlab}[1]{#1}

\bibitem[{Barron et~al.(2021)Barron, Mildenhall, Tancik, Hedman, Martin-Brualla, and Srinivasan}]{barron2021mip}
Barron, J.~T.; Mildenhall, B.; Tancik, M.; Hedman, P.; Martin-Brualla, R.; and Srinivasan, P.~P. 2021.
\newblock Mip-nerf: A multiscale representation for anti-aliasing neural radiance fields.
\newblock In \emph{Proceedings of the IEEE/CVF international conference on computer vision}, 5855--5864.

\bibitem[{Barron et~al.(2022)Barron, Mildenhall, Verbin, Srinivasan, and Hedman}]{barron2022mip}
Barron, J.~T.; Mildenhall, B.; Verbin, D.; Srinivasan, P.~P.; and Hedman, P. 2022.
\newblock Mip-nerf 360: Unbounded anti-aliased neural radiance fields.
\newblock In \emph{Proceedings of the IEEE/CVF conference on computer vision and pattern recognition}, 5470--5479.

\bibitem[{Bay, Tuytelaars, and Van~Gool(2006)}]{bay2006surf}
Bay, H.; Tuytelaars, T.; and Van~Gool, L. 2006.
\newblock Surf: Speeded up robust features.
\newblock In \emph{Computer Vision--ECCV 2006: 9th European Conference on Computer Vision, Graz, Austria, May 7-13, 2006. Proceedings, Part I 9}, 404--417. Springer.

\bibitem[{Calonder et~al.(2010)Calonder, Lepetit, Strecha, and Fua}]{calonder2010brief}
Calonder, M.; Lepetit, V.; Strecha, C.; and Fua, P. 2010.
\newblock Brief: Binary robust independent elementary features.
\newblock In \emph{Computer Vision--ECCV 2010: 11th European Conference on Computer Vision, Heraklion, Crete, Greece, September 5-11, 2010, Proceedings, Part IV 11}, 778--792. Springer.

\bibitem[{Chan et~al.(2021)Chan, Wang, Yu, Dong, and Loy}]{chan2021basicvsr}
Chan, K.~C.; Wang, X.; Yu, K.; Dong, C.; and Loy, C.~C. 2021.
\newblock Basicvsr: The search for essential components in video super-resolution and beyond.
\newblock In \emph{Proceedings of the IEEE/CVF conference on computer vision and pattern recognition}, 4947--4956.

\bibitem[{Chan et~al.(2022)Chan, Zhou, Xu, and Loy}]{chan2022basicvsr++}
Chan, K.~C.; Zhou, S.; Xu, X.; and Loy, C.~C. 2022.
\newblock Basicvsr++: Improving video super-resolution with enhanced propagation and alignment.
\newblock In \emph{Proceedings of the IEEE/CVF conference on computer vision and pattern recognition}, 5972--5981.

\bibitem[{Chen et~al.(2022)Chen, Xu, Geiger, Yu, and Su}]{chen2022tensorf}
Chen, A.; Xu, Z.; Geiger, A.; Yu, J.; and Su, H. 2022.
\newblock Tensorf: Tensorial radiance fields.
\newblock In \emph{European conference on computer vision}, 333--350. Springer.

\bibitem[{Chen et~al.(2023)Chen, Wang, Zhou, Qiao, and Dong}]{chen2023activating}
Chen, X.; Wang, X.; Zhou, J.; Qiao, Y.; and Dong, C. 2023.
\newblock Activating more pixels in image super-resolution transformer.
\newblock In \emph{Proceedings of the IEEE/CVF conference on computer vision and pattern recognition}, 22367--22377.

\bibitem[{Fan et~al.(2024)Fan, Yang, Li, Li, and Zhang}]{fan2024trim}
Fan, L.; Yang, Y.; Li, M.; Li, H.; and Zhang, Z. 2024.
\newblock Trim 3D Gaussian Splatting for Accurate Geometry Representation.
\newblock \emph{arXiv preprint arXiv:2406.07499}.

\bibitem[{Feng et~al.(2024{\natexlab{a}})Feng, He, Wang, Wang, Kuang, Ding, Qin, Yu, and Fan}]{feng2024zs}
Feng, X.; He, Y.; Wang, Y.; Wang, C.; Kuang, Z.; Ding, J.; Qin, F.; Yu, J.; and Fan, J. 2024{\natexlab{a}}.
\newblock ZS-SRT: An efficient zero-shot super-resolution training method for Neural Radiance Fields.
\newblock \emph{Neurocomputing}, 590: 127714.

\bibitem[{Feng et~al.(2024{\natexlab{b}})Feng, He, Wang, Yang, Kuang, Jun, Fan et~al.}]{feng2024srgs}
Feng, X.; He, Y.; Wang, Y.; Yang, Y.; Kuang, Z.; Jun, Y.; Fan, J.; et~al. 2024{\natexlab{b}}.
\newblock SRGS: Super-Resolution 3D Gaussian Splatting.
\newblock \emph{arXiv preprint arXiv:2404.10318}.

\bibitem[{Fridovich-Keil et~al.(2022)Fridovich-Keil, Yu, Tancik, Chen, Recht, and Kanazawa}]{fridovich2022plenoxels}
Fridovich-Keil, S.; Yu, A.; Tancik, M.; Chen, Q.; Recht, B.; and Kanazawa, A. 2022.
\newblock Plenoxels: Radiance fields without neural networks.
\newblock In \emph{Proceedings of the IEEE/CVF conference on computer vision and pattern recognition}, 5501--5510.

\bibitem[{Gu{\'e}don and Lepetit(2024)}]{guedon2024sugar}
Gu{\'e}don, A.; and Lepetit, V. 2024.
\newblock Sugar: Surface-aligned gaussian splatting for efficient 3d mesh reconstruction and high-quality mesh rendering.
\newblock In \emph{Proceedings of the IEEE/CVF Conference on Computer Vision and Pattern Recognition}, 5354--5363.

\bibitem[{Han et~al.(2023)Han, Yu, Yu, Wang, and Dai}]{Han2023super}
Han, Y.; Yu, T.; Yu, X.; Wang, Y.; and Dai, Q. 2023.
\newblock Super-NeRF: View-consistent Detail Generation for NeRF super-resolution.
\newblock \emph{arXiv preprint arXiv:2304.13518}.

\bibitem[{Huang et~al.(2024)Huang, Yu, Chen, Geiger, and Gao}]{huang20242d}
Huang, B.; Yu, Z.; Chen, A.; Geiger, A.; and Gao, S. 2024.
\newblock 2d gaussian splatting for geometrically accurate radiance fields.
\newblock In \emph{ACM SIGGRAPH 2024 Conference Papers}, 1--11.

\bibitem[{Kerbl et~al.(2023)Kerbl, Kopanas, Leimk{\"u}hler, and Drettakis}]{kerbl20233d}
Kerbl, B.; Kopanas, G.; Leimk{\"u}hler, T.; and Drettakis, G. 2023.
\newblock 3D Gaussian Splatting for Real-Time Radiance Field Rendering.
\newblock \emph{ACM Trans. Graph.}, 42(4): 139--1.

\bibitem[{Lee, Li, and Lee(2024)}]{lee2024disr}
Lee, J.~L.; Li, C.; and Lee, G.~H. 2024.
\newblock DiSR-NeRF: Diffusion-Guided View-Consistent Super-Resolution NeRF.
\newblock In \emph{Proceedings of the IEEE/CVF Conference on Computer Vision and Pattern Recognition}, 20561--20570.

\bibitem[{Liang et~al.(2024)Liang, Cao, Fan, Zhang, Ranjan, Li, Timofte, and Van~Gool}]{liang2024vrt}
Liang, J.; Cao, J.; Fan, Y.; Zhang, K.; Ranjan, R.; Li, Y.; Timofte, R.; and Van~Gool, L. 2024.
\newblock Vrt: A video restoration transformer.
\newblock \emph{IEEE Transactions on Image Processing}.

\bibitem[{Liang et~al.(2021)Liang, Cao, Sun, Zhang, Van~Gool, and Timofte}]{liang2021swinir}
Liang, J.; Cao, J.; Sun, G.; Zhang, K.; Van~Gool, L.; and Timofte, R. 2021.
\newblock Swinir: Image restoration using swin transformer.
\newblock In \emph{Proceedings of the IEEE/CVF international conference on computer vision}, 1833--1844.

\bibitem[{Lim et~al.(2017)Lim, Son, Kim, Nah, and Mu~Lee}]{lim2017enhanced}
Lim, B.; Son, S.; Kim, H.; Nah, S.; and Mu~Lee, K. 2017.
\newblock Enhanced deep residual networks for single image super-resolution.
\newblock In \emph{Proceedings of the IEEE conference on computer vision and pattern recognition workshops}, 136--144.

\bibitem[{Lin et~al.(2024)Lin, Fu, Merth, Yang, and Ranjan}]{lin2024fastsr}
Lin, C.-Y.; Fu, Q.; Merth, T.; Yang, K.; and Ranjan, A. 2024.
\newblock Fastsr-nerf: Improving nerf efficiency on consumer devices with a simple super-resolution pipeline.
\newblock In \emph{Proceedings of the IEEE/CVF Winter Conference on Applications of Computer Vision}, 6036--6045.

\bibitem[{Liu et~al.(2023)Liu, Wu, Van~Hoorick, Tokmakov, Zakharov, and Vondrick}]{liu2023zero}
Liu, R.; Wu, R.; Van~Hoorick, B.; Tokmakov, P.; Zakharov, S.; and Vondrick, C. 2023.
\newblock Zero-1-to-3: Zero-shot one image to 3d object.
\newblock In \emph{Proceedings of the IEEE/CVF international conference on computer vision}, 9298--9309.

\bibitem[{Lowe(2004)}]{lowe2004distinctive}
Lowe, D.~G. 2004.
\newblock Distinctive image features from scale-invariant keypoints.
\newblock \emph{International journal of computer vision}, 60: 91--110.

\bibitem[{Mildenhall et~al.(2021)Mildenhall, Srinivasan, Tancik, Barron, Ramamoorthi, and Ng}]{mildenhall2021nerf}
Mildenhall, B.; Srinivasan, P.~P.; Tancik, M.; Barron, J.~T.; Ramamoorthi, R.; and Ng, R. 2021.
\newblock Nerf: Representing scenes as neural radiance fields for view synthesis.
\newblock \emph{Communications of the ACM}, 65(1): 99--106.

\bibitem[{M{\"u}ller et~al.(2022)M{\"u}ller, Evans, Schied, and Keller}]{muller2022instant}
M{\"u}ller, T.; Evans, A.; Schied, C.; and Keller, A. 2022.
\newblock Instant neural graphics primitives with a multiresolution hash encoding.
\newblock \emph{ACM transactions on graphics (TOG)}, 41(4): 1--15.

\bibitem[{Poole et~al.(2022)Poole, Jain, Barron, and Mildenhall}]{poole2022dreamfusion}
Poole, B.; Jain, A.; Barron, J.~T.; and Mildenhall, B. 2022.
\newblock Dreamfusion: Text-to-3d using 2d diffusion.
\newblock \emph{arXiv preprint arXiv:2209.14988}.

\bibitem[{Rombach et~al.(2022)Rombach, Blattmann, Lorenz, Esser, and Ommer}]{rombach2022high}
Rombach, R.; Blattmann, A.; Lorenz, D.; Esser, P.; and Ommer, B. 2022.
\newblock High-resolution image synthesis with latent diffusion models.
\newblock In \emph{Proceedings of the IEEE/CVF conference on computer vision and pattern recognition}, 10684--10695.

\bibitem[{Rosten and Drummond(2006)}]{rosten2006machine}
Rosten, E.; and Drummond, T. 2006.
\newblock Machine learning for high-speed corner detection.
\newblock In \emph{Computer Vision--ECCV 2006: 9th European Conference on Computer Vision, Graz, Austria, May 7-13, 2006. Proceedings, Part I 9}, 430--443. Springer.

\bibitem[{Rublee et~al.(2011)Rublee, Rabaud, Konolige, and Bradski}]{rublee2011orb}
Rublee, E.; Rabaud, V.; Konolige, K.; and Bradski, G. 2011.
\newblock ORB: An efficient alternative to SIFT or SURF.
\newblock In \emph{Proceedings of the IEEE International Conference on Computer Vision}, 2564--2571. IEEE.

\bibitem[{Shen et~al.(2024)Shen, Ceylan, Guerrero, Xu, Mitra, Wang, and Fr{\"u}st{\"u}ck}]{shen2024supergaussian}
Shen, Y.; Ceylan, D.; Guerrero, P.; Xu, Z.; Mitra, N.~J.; Wang, S.; and Fr{\"u}st{\"u}ck, A. 2024.
\newblock SuperGaussian: Repurposing Video Models for 3D Super Resolution.
\newblock \emph{arXiv preprint arXiv:2406.00609}.

\bibitem[{Shi et~al.(2022)Shi, Gu, Xie, Wang, Yang, and Dong}]{shi2022rethinking}
Shi, S.; Gu, J.; Xie, L.; Wang, X.; Yang, Y.; and Dong, C. 2022.
\newblock Rethinking alignment in video super-resolution transformers.
\newblock \emph{Advances in Neural Information Processing Systems}, 35: 36081--36093.

\bibitem[{Tian et~al.(2024)Tian, Chen, Xu, and Wang}]{tian2024image}
Tian, Y.; Chen, H.; Xu, C.; and Wang, Y. 2024.
\newblock Image Processing GNN: Breaking Rigidity in Super-Resolution.
\newblock In \emph{Proceedings of the IEEE/CVF Conference on Computer Vision and Pattern Recognition}, 24108--24117.

\bibitem[{Voleti et~al.(2024)Voleti, Yao, Boss, Letts, Pankratz, Tochilkin, Laforte, Rombach, and Jampani}]{voleti2024sv3d}
Voleti, V.; Yao, C.-H.; Boss, M.; Letts, A.; Pankratz, D.; Tochilkin, D.; Laforte, C.; Rombach, R.; and Jampani, V. 2024.
\newblock Sv3d: Novel multi-view synthesis and 3d generation from a single image using latent video diffusion.
\newblock \emph{arXiv preprint arXiv:2403.12008}.

\bibitem[{Wang et~al.(2022)Wang, Wu, Guo, Zhang, Tai, and Hu}]{wang2022nerf}
Wang, C.; Wu, X.; Guo, Y.-C.; Zhang, S.-H.; Tai, Y.-W.; and Hu, S.-M. 2022.
\newblock Nerf-sr: High quality neural radiance fields using supersampling.
\newblock In \emph{Proceedings of the 30th ACM International Conference on Multimedia}, 6445--6454.

\bibitem[{Wang et~al.(2021)Wang, Liu, Liu, Theobalt, Komura, and Wang}]{wang2021neus}
Wang, P.; Liu, L.; Liu, Y.; Theobalt, C.; Komura, T.; and Wang, W. 2021.
\newblock Neus: Learning neural implicit surfaces by volume rendering for multi-view reconstruction.
\newblock \emph{arXiv preprint arXiv:2106.10689}.

\bibitem[{Wang et~al.(2024)Wang, Leroy, Cabon, Chidlovskii, and Revaud}]{wang2024dust3r}
Wang, S.; Leroy, V.; Cabon, Y.; Chidlovskii, B.; and Revaud, J. 2024.
\newblock Dust3r: Geometric 3d vision made easy.
\newblock In \emph{Proceedings of the IEEE/CVF Conference on Computer Vision and Pattern Recognition}, 20697--20709.

\bibitem[{Wang et~al.(2018)Wang, Yu, Wu, Gu, Liu, Dong, Qiao, and Change~Loy}]{wang2018esrgan}
Wang, X.; Yu, K.; Wu, S.; Gu, J.; Liu, Y.; Dong, C.; Qiao, Y.; and Change~Loy, C. 2018.
\newblock Esrgan: Enhanced super-resolution generative adversarial networks.
\newblock In \emph{Proceedings of the European Conference on Computer Vision (ECCV) Workshops}, 0--0.

\bibitem[{Wu et~al.(2024)Wu, Wan, Zhang, Liao, and Xu}]{wu2024rafe}
Wu, Z.; Wan, Z.; Zhang, J.; Liao, J.; and Xu, D. 2024.
\newblock RaFE: Generative Radiance Fields Restoration.
\newblock \emph{arXiv preprint arXiv:2404.03654}.

\bibitem[{Xie et~al.(2024)Xie, Wang, Zhu, and Pan}]{xie2024supergs}
Xie, S.; Wang, Z.; Zhu, Y.; and Pan, C. 2024.
\newblock SuperGS: Super-Resolution 3D Gaussian Splatting via Latent Feature Field and Gradient-guided Splitting.
\newblock \emph{arXiv preprint arXiv:2410.02571}.

\bibitem[{Xu et~al.(2024{\natexlab{a}})Xu, Yu, Wang, Mi, and Yao}]{xu2024enhancing}
Xu, K.; Yu, Z.; Wang, X.; Mi, M.~B.; and Yao, A. 2024{\natexlab{a}}.
\newblock Enhancing Video Super-Resolution via Implicit Resampling-based Alignment.
\newblock In \emph{Proceedings of the IEEE/CVF Conference on Computer Vision and Pattern Recognition}, 2546--2555.

\bibitem[{Xu et~al.(2024{\natexlab{b}})Xu, Park, Zhang, Zhou, Shechtman, Liu, Huang, and Liu}]{xu2024videogigagan}
Xu, Y.; Park, T.; Zhang, R.; Zhou, Y.; Shechtman, E.; Liu, F.; Huang, J.-B.; and Liu, D. 2024{\natexlab{b}}.
\newblock VideoGigaGAN: Towards Detail-rich Video Super-Resolution.
\newblock \emph{arXiv preprint arXiv:2404.12388}.

\bibitem[{Yariv et~al.(2021)Yariv, Gu, Kasten, and Lipman}]{yariv2021volume}
Yariv, L.; Gu, J.; Kasten, Y.; and Lipman, Y. 2021.
\newblock Volume rendering of neural implicit surfaces.
\newblock \emph{Advances in Neural Information Processing Systems}, 34: 4805--4815.

\bibitem[{Yariv et~al.(2023)Yariv, Hedman, Reiser, Verbin, Srinivasan, Szeliski, Barron, and Mildenhall}]{yariv2023bakedsdf}
Yariv, L.; Hedman, P.; Reiser, C.; Verbin, D.; Srinivasan, P.~P.; Szeliski, R.; Barron, J.~T.; and Mildenhall, B. 2023.
\newblock Bakedsdf: Meshing neural sdfs for real-time view synthesis.
\newblock In \emph{ACM SIGGRAPH 2023 Conference Proceedings}, 1--9.

\bibitem[{Yoon and Yoon(2023)}]{yoon2023cross}
Yoon, Y.; and Yoon, K.-J. 2023.
\newblock Cross-guided optimization of radiance fields with multi-view image super-resolution for high-resolution novel view synthesis.
\newblock In \emph{Proceedings of the IEEE/CVF Conference on Computer Vision and Pattern Recognition}, 12428--12438.

\bibitem[{Yu et~al.(2024{\natexlab{a}})Yu, Zhu, He, and Chen}]{yu2024gaussiansr}
Yu, X.; Zhu, H.; He, T.; and Chen, Z. 2024{\natexlab{a}}.
\newblock GaussianSR: 3D Gaussian Super-Resolution with 2D Diffusion Priors.
\newblock \emph{arXiv preprint arXiv:2406.10111}.

\bibitem[{Yu et~al.(2024{\natexlab{b}})Yu, Chen, Huang, Sattler, and Geiger}]{yu2024mip}
Yu, Z.; Chen, A.; Huang, B.; Sattler, T.; and Geiger, A. 2024{\natexlab{b}}.
\newblock Mip-splatting: Alias-free 3d gaussian splatting.
\newblock In \emph{Proceedings of the IEEE/CVF Conference on Computer Vision and Pattern Recognition}, 19447--19456.

\bibitem[{Zhang et~al.(2021)Zhang, Liang, Van~Gool, and Timofte}]{zhang2021designing}
Zhang, K.; Liang, J.; Van~Gool, L.; and Timofte, R. 2021.
\newblock Designing a practical degradation model for deep blind image super-resolution.
\newblock In \emph{Proceedings of the IEEE/CVF International Conference on Computer Vision}, 4791--4800.

\bibitem[{Zuo et~al.(2024)Zuo, Gu, Qiu, Dong, Zhao, Yuan, Peng, Zhu, Dong, Bo et~al.}]{zuo2024videomv}
Zuo, Q.; Gu, X.; Qiu, L.; Dong, Y.; Zhao, Z.; Yuan, W.; Peng, R.; Zhu, S.; Dong, Z.; Bo, L.; et~al. 2024.
\newblock Videomv: Consistent multi-view generation based on large video generative model.
\newblock \emph{arXiv preprint arXiv:2403.12010}.

\end{thebibliography}

\newpage
\appendix

\section*{Appendix}\label{sec:reference_examples}

\begin{table*}[ht]
    \centering
    \caption{Ablation comparison of Blender dataset ($\times4 \rightarrow \times1$) on various VSR models. SISR refers to Single-Image Super-Resolution (single image VSR), S refers to ordering by simple greedy algorithm (order: feature), and ALS refers to using adaptive-length subsequence (order: feature) with multi-threshold (threshold: pose).}
    \label{tab:VSR_baselines_blender}
    \begin{tabular}{|c||c c c||c c c||c c c|}
    \hline
       & \multicolumn{3}{c||}{VRT} & \multicolumn{3}{c||}{IART} & \multicolumn{3}{c|}{PSRT} \\
     \hline
       & PSNR$\uparrow$ & SSIM$\uparrow$ & LPIPS$\downarrow$ & PSNR$\uparrow$ & SSIM$\uparrow$ & LPIPS$\downarrow$ & PSNR$\uparrow$ & SSIM$\uparrow$ & LPIPS$\downarrow$ \\
    \hline

      SISR & 31.20   & 0.9497 & 0.0567 & 31.10 & 0.9484 & 0.0590 & 31.10 & 0.9516 & 0.0543 \\
      S    & 31.25  & 0.9505 & 0.0557 & 31.32 & 0.9513 & 0.0550 & 31.35 & 0.9513 & 0.0548 \\
      ALS & 31.37   & 0.9516 & 0.0544  & 31.35 & 0.9514 & 0.0548 & 31.41 & 0.9520 & 0.0540 \\
    \hline
    \end{tabular}
\end{table*}

\section{Flexibility with VSR Baseline Models}

One of the key advantages of our model is its ability to utilize any pre-trained VSR model as a backbone, unlike SuperGaussian, which necessitates extensive and costly training to align the pre-trained data distribution with that of Gaussian splats. We demonstrate the robustness of our approach by integrating various pre-trained VSR models. Specifically, we evaluated three VSR models—VRT, PSRT, and IART—alongside ablation studies on the simple greedy algorithm (S) and adaptive-length subsequence generation (ALS), comparing these with the original unordered sequence and SISR (where the VSR model processes each image individually without reference frames). Tab.~\ref{tab:VSR_baselines_blender} demonstrate that our method is flexible with the choice of VSR models.

\section{Implementation Details}
We implement our method using the open-source 3D Gaussian Splatting code base. Following the 3DGS protocol, we train the 3DGS model for 30,000 iterations. To create the low-resolution (LR) dataset, we downsample the high-resolution (HR) dataset using bicubic interpolation with a downscale factor of 4.

To evaluate the generalization capabilities of Video Super-Resolution (VSR) models within our framework, we conduct ablation experiments on three VSR models: VRT, IART, and PSRT. By default, these models are pre-trained on the LR Vimeo-90K dataset, which is downsampled using bicubic interpolation. 

For the synthetic Blender dataset, we utilize nearest neighbor ordering based on ORB features and apply thresholds based on pose similarity. In contrast, for the Mip-NeRF 360 dataset, we employ nearest neighbor matching based on pose and apply thresholds using ORB features. This approach is justified by the object-centric nature of the Blender dataset and the non-object-centric characteristics of the Mip-NeRF 360 dataset.

For the Adaptive-length Sequences, we set the three thresholds (angle between two camera positions) to $15^{\circ}$, $30^{\circ}$, $45^{\circ}$ on both Blender dataset. We set two thresholds (the number of candidates by distances) to 30 and 50.

\begin{figure}[t]
\centering
\includegraphics[width=0.95\columnwidth]{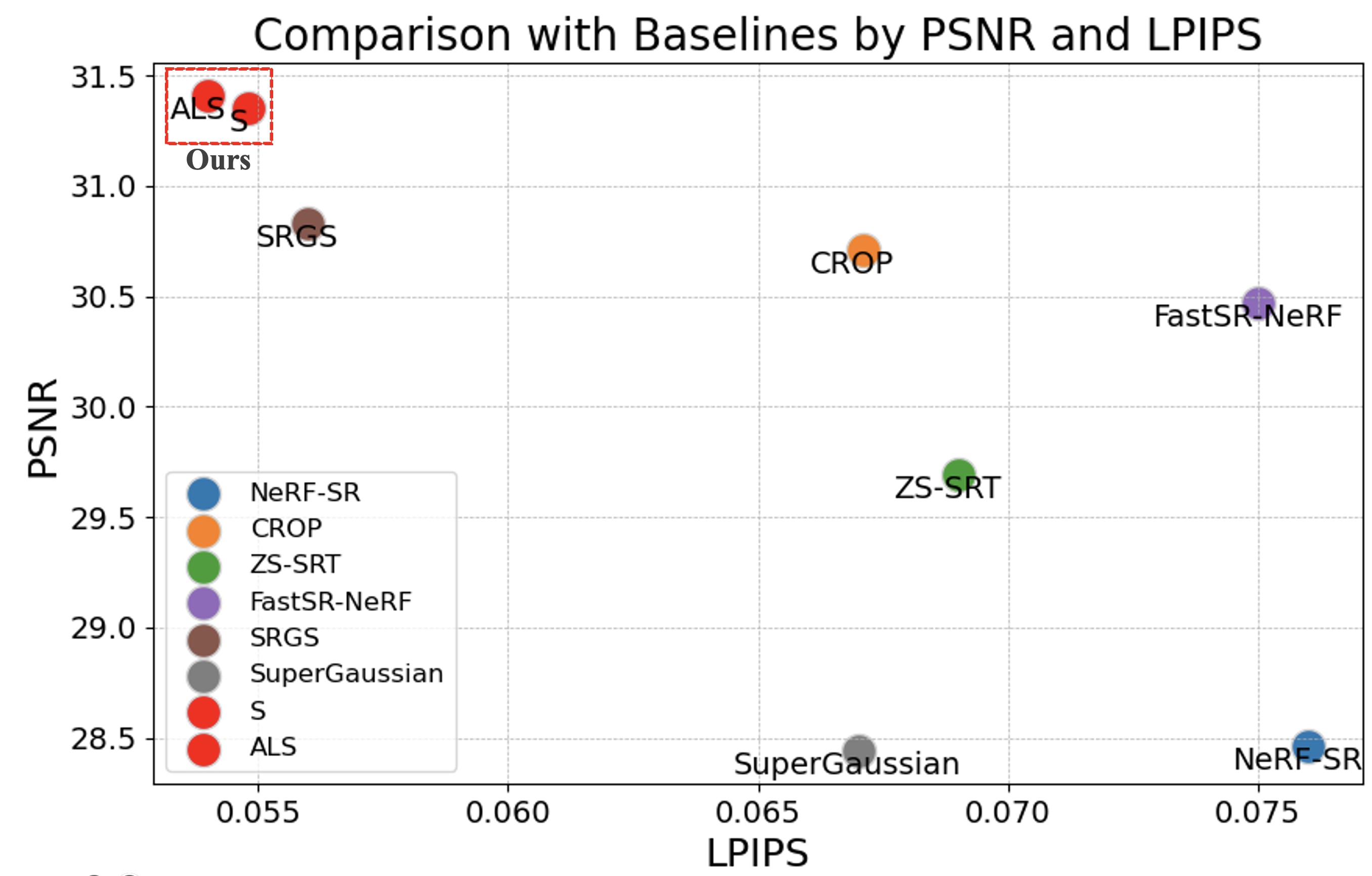}
\caption{Comparison with baselines.}
\label{fig:baseline}
\end{figure}

\begin{figure}[t]
\centering
\includegraphics[width=\columnwidth]{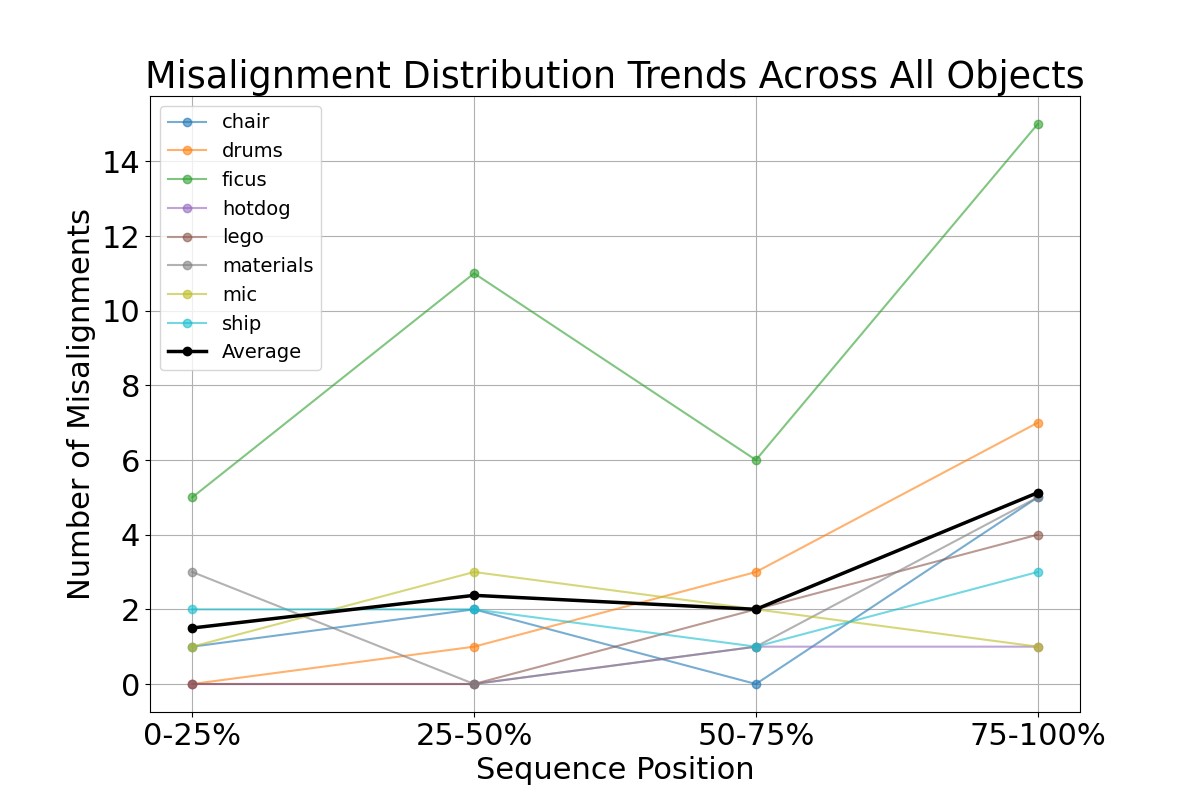} % Reduce the figure size so that it is slightly narrower than the column.
\caption{Misalignment trends within a sequence.}
\label{fig9}
\end{figure}

\section{Misalignment Error}

\begin{figure}[ht!]
\centering
\includegraphics[width=0.95\columnwidth]{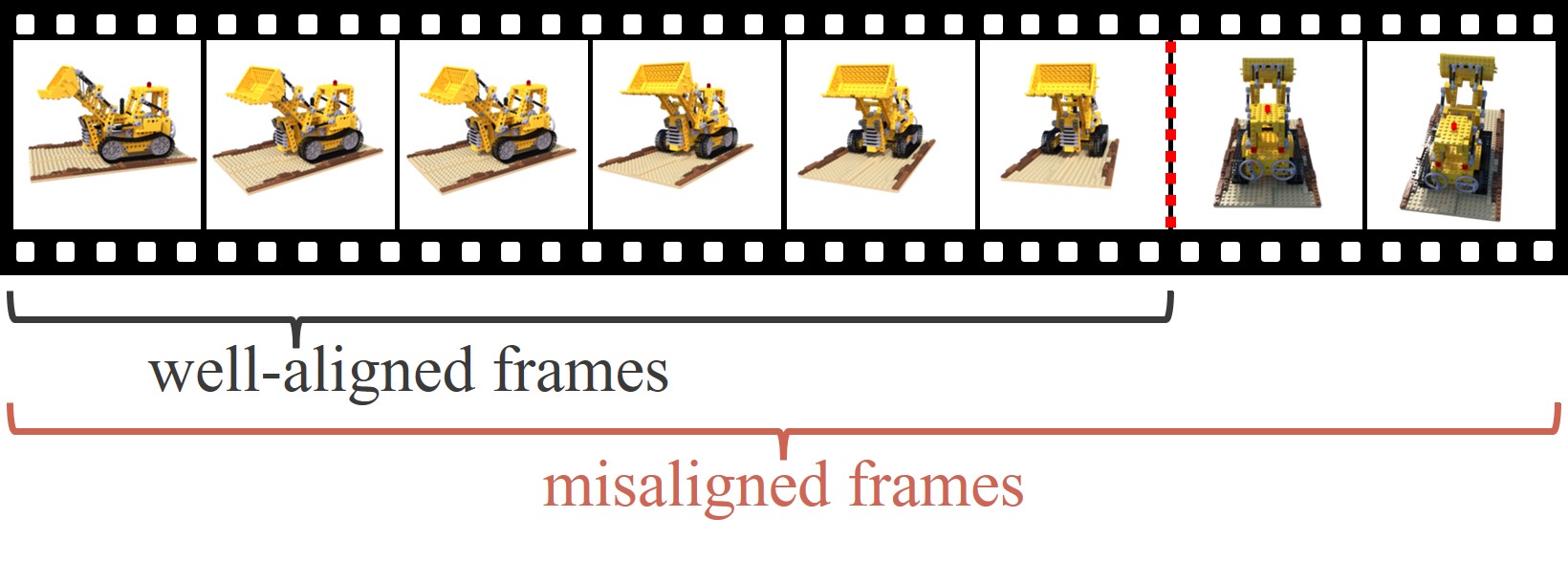} % Reduce the figure size so that it is slightly narrower than the column.
\caption{Misalignment Error.}
\label{fig:misalignment}
\end{figure}

\begin{table}[ht]
    \centering
    \caption{Impact of misalignment on 3D super-resolution. }
    \label{tab:impact_of_misalignment}
    \begin{tabular}{|c||c|c|c|}
        \hline
            & PSNR$\uparrow$ & SSIM$\uparrow$ & LPIPS$\downarrow$ \\
        \hline
        S (last 25\%)  & 31.32 & 0.9511 &  0.0552 \\
        ALS & 31.41 & 0.9520 & 0.0540 \\
        \hline
    \end{tabular}
\end{table}

Misalignment errors (Fig.~\ref{fig:misalignment}) occur due to inaccuracies in aligning frames, particularly when ORB features are used to link them. As the number of frames in a sequence increases, the probability of connecting unrelated features—especially from temporally distant frames—also rises, as images that have been connected once cannot be reconnected. This misalignment may cause the model to rely on incorrect or irrelevant frame information during upsampling, thereby compromising the quality of the output.

Our analysis reveals that misalignment errors escalate as the sequence length increases, since longer sequences provide more opportunities for feature mismatches. To quantify this, we first extract the camera's center position and direction (z-axis of the camera coordinates) in world coordinates from the transformation matrix. We classify a frame as misaligned if the consecutive images exhibit an angular difference greater than $45^{\circ}$, when measured between the vectors drawn from the camera positions to the origin.

To demonstrate the impact of misalignments, we conducted a toy experiment. As illustrated in Fig.~\ref{fig9}, misalignments tend to increase towards the end of the sequence, where unconnected images are forcefully connected, which significantly degrades the performance of our simple greedy algorithm. To quantify this degradation, we focus on the last 25\% of the sequence generated by the greedy algorithm. To construct a complete sequence, we apply our greedy algorithm starting from each image, collecting the last 25\% of each sequence. These segments are then combined to form the final sequence, which is expected to exhibit a high degree of misalignment. For any images that could not be included in the final sequence using this method, we directly use the upsampled images generated by the ALS (adaptive-length subsequence). This approach allows us to highlight the misalignment issues inherent in the greedy algorithm, in comparison to ALS (adaptive-length subsequence). The results are shown in Tab.~\ref{tab:impact_of_misalignment}.

\section{Per-object and Per-scene Quantitative Results}
We present per-object (synthetic Blender) and per-scene (Mip-NeRF 360) PSNR comparisons with different baseline models. Our method uses PSRT as our VSR backbone using adaptive-length subsequence (ALS) with multi-threshold. Note that value marked with $\dagger$ is taken from the respective paper, as the code for the model is not available. The results can be found in Tab. \ref{tab:per-object} and Tab. \ref{tab:per-scene}.

\begin{table*}[t!]
    \caption{Per-object PSNR comparison on the synthetic Blender dataset ($\times4$ $\rightarrow$ $\times1$). Ours-ALS refers to our method using adaptive-length subsequencing (ALS).}
    \label{tab:per-object}
    \centering
    \begin{tabular}{|l|cccccccc|c|}
    \hline
    & chair & drums & ficus & hotdog & lego & materials & mic & ship & average \\
    \hline
    Bicubic                 &29.02 &23.75 &28.24 &31.86 &27.46 &26.47 &27.97 &25.71 & 27.56   \\
    PSRT (SISR) & 30.94 & 25.56 & 33.49 & 35.82 & 32.20 & 30.06 & 31.75 & 28.96 & 31.10 \\
    SwinIR+3DGS   &31.02 &25.48 &32.49 &35.60 &32.05 &29.58 &31.75 &28.20 &30.77   \\
    Render-SR       &30.23 &24.04 &28.63 &33.78 &29.23 &27.34 &30.53 &27.35 & 28.90   \\
    NeRF-SR       &30.16 &23.46 &26.64 &34.40 &29.13 &28.02 &27.25 &26.61 & 28.21   \\
    DiSR-NeRF    & 27.55 & 22.63 & 25.64 & 30.07 & 26.43 & 24.71 & 26.49 & 24.47 & 26.00  \\
    CROP$^{\dagger}$    &31.53 &24.99 &31.50 &35.62 &32.88 &29.16 &31.76 &28.23 & 30.71  \\
    Ours-S    &31.33 &25.58 &33.71 &35.95 &32.98 &30.09 &31.91 &29.26 &31.35 \\
    Ours-ALS    & 31.36 &25.65 &33.69 &36.18 &33.03 &30.17 &31.93 &29.26 &31.41 \\
    \hline
    HR-3DGS    &35.79 &26.14 &34.84 &37.72 &35.77 &29.97 &35.36 &30.89 &33.31 \\
    \hline
    \end{tabular}
\end{table*}

\begin{table*}[t!]
    \caption{Per-object SSIM comparison on the synthetic Blender dataset ($\times4$ $\rightarrow$ $\times1$). Ours-ALS refers to our method using adaptive-length subsequencing (ALS).}
    \label{tab:per-object_lpips}
    \centering
    \begin{tabular}{|l|cccccccc|c|}
    \hline
    & chair & drums & ficus & hotdog & lego & materials & mic & ship & average \\
    \hline
    Bicubic        &0.9194 &0.9003 &0.9430 &0.9526 &0.9059 &0.9220 &0.9481 &0.8291 &0.9150 \\

    PSRT (SISR) & 0.9475 & 0.9386 & 0.9762 & 0.9721 & 0.9572 & 0.9544 & 0.9732 & 0.8688 & 0.9516 \\
    SwinIR+3DGS   &0.9469 &0.9412 &0.9760 &0.9728 &0.9601 &0.9558 &0.9747 &0.8731 &0.9501    \\
    Render-SR       &0.9432 &0.9163 &0.9539 &0.9677 &0.9379 &0.9322 &0.9671 &0.8582 &0.9346    \\
    NeRF-SR       &0.9366 &0.9019 &0.9026 &0.9629 &0.9292 &0.9319 &0.9432 &0.8357 &0.9180    \\
    DiSR-NeRF    &0.9035 &0.8618 &0.9117 &0.9332 &0.8875 &0.8816 &0.9335 &0.8053 &0.8898    \\
    CROP$^{\dagger}$    &0.9513 &0.9236 &0.9709 & 0.9725 &0.9641 &0.9468 & 0.9740 &0.8637 &0.9459    \\
    Ours-S    & 0.9538 & 0.9391 & 0.9779 & 0.9738 & 0.9646 & 0.9541 & 0.9747 & 0.8724 & 0.9513 \\
    Ours-ALS    & 0.9539 & 0.9405 & 0.9777 & 0.9744 & 0.9649 & 0.9555 & 0.9750 & 0.8741 & 0.9520 \\
    \hline
    HR-3DGS    &0.9874 &0.9544 &0.9872 &0.9853 &0.9828 &0.9603 &0.9914 &0.9067 &0.9694    \\
    \hline
    \end{tabular}
\end{table*}

\begin{table*}[t!]
    \caption{Per-object LPIPS comparison on the synthetic Blender dataset ($\times4$ $\rightarrow$ $\times1$). Ours-ALS refers to our method using adaptive-length subsequencing (ALS).}
    \label{tab:per-object_lpips}
    \centering
    \begin{tabular}{|l|cccccccc|c|}
    \hline
    & chair & drums & ficus & hotdog & lego & materials & mic & ship & average \\
    \hline
    Bicubic  & 0.0899 & 0.1106 & 0.0619 & 0.0768 & 0.1272 & 0.0892 & 0.0626 & 0.2136 &0.1040  \\

    PSRT (SISR) & 0.0553 &0.0609 & 0.0237&0.0421 &0.0595 &0.0480 & 0.0254 &0.1567 & 0.0544  \\
    SwinIR+3DGS   & 0.0577 & 0.0565 & 0.0221 & 0.0401 &0.0498 & 0.0420 & 0.0203 & 0.1511 & 0.0550   \\
    Render-SR & 0.0563 & 0.0743 & 0.0396 & 0.0462 & 0.0691 & 0.0597 & 0.0312 & 0.1698 & 0.0683  \\
    NeRF-SR       & 0.0687 & 0.1091 & 0.1014 & 0.0591 & 0.0976 & 0.0770 & 0.0805 & 0.1984 & 0.0990   \\
    DiSR-NeRF    &0.0943 & 0.1429 &0.0905 &0.1001 &0.1378 &0.1293 & 0.0751&0.2106 &0.1226    \\
    CROP$^{\dagger}$    &0.0567 &0.0856 &0.0317 &0.0481 &0.0496 &0.0622 &0.0251 &0.1776 &0.0671    \\
    Ours-S    & 0.0478 & 0.0585 & 0.0216 & 0.0395 & 0.0470 & 0.0488 & 0.0240 & 0.1509 & 0.0547 \\
    Ours-ALS    & 0.0478 & 0.0576 & 0.0216 & 0.0388 & 0.0465 & 0.0464 & 0.0233 & 0.1501 & 0.0540 \\
    \hline
    HR-3DGS  &0.0117 &0.0371 &0.0116 &0.0199 &0.0154 &0.0341 &0.0060 &0.1063 &0.0303    \\
    \hline
    \end{tabular}
\end{table*}

\section{Multi-threshold Subsequence}
\label{sec:multi_threshold_subsequence}
In the main paper, we introduced the concept of multi-threshold subsequence generation. To summarize briefly, applying a uniform threshold across all sequences can be inefficient due to varying image densities. A strict threshold ensures that only closely situated images are connected, resulting in a smoother trajectory. However, in sparsely populated regions, images are rarely connected with a strict threshold, leading to a loss of reference when upsampling. Conversely, a loose threshold connects images even over greater distances, ensuring that most images are connected, but potentially sacrificing smoothness in densely populated regions. To address this, we propose a multi-threshold subsequence generation method. We first upsample images using a strict threshold to benefit from smoother trajectories in dense regions. Then, we gradually loosen the threshold to generate less smooth trajectories; this way, we can ensure that most images achieve the smoothest trajectory possible.

\section{Sub-pixel Loss and Final Loss}
In this section, we will provide a detailed explanation of sub-pixel loss and final loss through equations. Let $\hat{I}, \tilde{I}, I \in \mathbb{R}^{H \times W \times 3}$ denotes a rendered image from 3DGS, the upsampled image via VSR models, and the ground-truth image, respectively. $H,W$ refers to the height and width of the HR images (we omitted the image index for brevity). According to 3DGS, the objective is written as below,
\begin{equation}
\label{3dgs_loss}
    \mathcal{L}_{\text{ren}} = (1-\lambda_1)\mathcal{L}_{1}(\hat{I}, \tilde{I}) + \lambda_1\mathcal{L}_{\mathrm{D-SSIM}}(\hat{I}, \tilde{I}).
\end{equation}
We use $\lambda_1=0.2$ in all our experiments. $\mathcal{L}_{1}(\cdot, \cdot)$ is L1 loss and $\mathcal{L}_{\mathrm{D-SSIM}}(\cdot, \cdot)$ is defined as $1 - \mathrm{SSIM}(\cdot, \cdot)$.

\begin{equation}
\label{sp_loss}
    \mathcal{L}_{sp} = (1-\lambda_1)\mathcal{L}_{1}(\downarrow(\hat{I}), \downarrow(I)) + \lambda_1\mathcal{L}_{\mathrm{D-SSIM}}(\downarrow(\hat{I}), \downarrow(I)),
\end{equation}
where $\downarrow(\cdot)$ is bicubic downsampling. And the final loss is defined as,
\begin{equation}
\label{final_loss}
    \mathcal{L} = \lambda_\text{ren} \mathcal{L}_\text{ren} + (1-\lambda_\text{ren}) \mathcal{L}_{sp}.
\end{equation}
We use $\lambda_\text{ren}=0.6$ for Blender dataset and $\lambda_\text{ren}=0.4$ for Mip-NeRF 360 dataset.

\section{ORB Feature Matching}
\label{appendix:ORB_feature_matching}
In the main paper, we discussed how ORB features are suitable for ordering unordered multi-view images into video sequences. In this section, we provide a detailed explanation of computing similarity scores using the ORB feature.

The similarity score \( \text{sim}(\cdot, \cdot) \) between two images \( I_i \) and \( I_j \) is computed as follows,
\[
S(I_i, I_j) = \frac{1}{|M(I_i,I_j)|} \sum_{(k, l) \in M(I_i,I_j)} \operatorname{dist}(f_{i,k}, f_{j,l}),
\]
where \( M(I_i,I_j) \) is a set of indices for matched descriptors between \( I_i \) and \( I_j \), $f_i$ is the ORB feature extracted from the image $I_i$, and $f_{i,k} \in \{0, 1\}^P$ is a binary feature vector for $k$-th keypoint in the image $I_i$, and $P=256$. The Hamming distance \( \operatorname{dist}(\cdot, \cdot) \) between two binary descriptors \( f_{i,k} \) and \( f_{j,l} \) is calculated as follows,
\[
\operatorname{dist}(f_{i,k},f_{j,l}) = \sum_{b=1}^{P} \left( f_{i,k,b} \oplus f_{j,l,b} \right),
\]
where \( f_{i,k,b} \in \{0, 1\} \) denotes \( b \)-th bits of descriptors \( f_{i,k} \), and \( \oplus \) is XOR operator.

Descriptors between images \( I_i \) and \( I_j \) are then matched using a bidirectional matching approach, also known as cross-checking. This process ensures robust matching by retaining only mutual best matches. Specifically, for each descriptor \( f_{i,k} \) in image \( I_i \), the descriptor \( f_{j,l} \) in image \( I_j \) with the smallest Hamming distance is identified, and vice versa. Only pairs \( (f_{i,k}, f_{j,l}) \) that are mutual best matches are retained. The set of indices of these matched descriptor pairs is denoted as \( M(I_i,I_j) \).

\section{Temporal Consistency}
\label{appendix:ORB_feature_matching}
Temporal consistency is crucial for our task, as VSR models rely on the coherence of neighboring frames to achieve better performance. By leveraging this temporal relationship, our method ensures 3D spatial consistency improving the 3D reconstruction quality. To evaluate temporal coherence, we use the Fréchet Video Distance (FVD) metric on the Blender dataset, where smooth video trajectories from the test split serve as ground truth. As shown in Tab.~\ref{tab:temporal_consistency}, our method (Ours-S and Ours-ALS) achieves the lowest FVD scores among all compared methods, demonstrating superior temporal consistency in video metrics. This improvement is attributed to the structured 'video-like' sequences generated by our ordering algorithms, which enhance both frame-to-frame coherence and spatial reconstruction accuracy.

\begin{table*}[t!]
    \caption{Per-scene PSNR comparison on the Mip-NeRF 360 dataset ($\times8$ $\rightarrow $$\times2$). Ours-ALS refers to our method using adaptive-length subsequencing (ALS).}
    \label{tab:per-scene}
    \centering
    \begin{tabular}{|l|ccccc|cccc|c|}
    \hline
    & bicycle & flowers & garden & stump & treehill & room & counter & kitchen & bonsai & average \\
    \hline
    Bicubic                 &24.02 &21.24 &25.14 &26.30 &22.25 &30.47 &28.15 &28.23 &30.21 &26.22  \\
    SwinIR + 3DGS   &24.54 &21.18 &25.81 &26.38 &22.16 &31.30 &28.71 &29.82 &31.26 &26.80  \\
    Ours-S       &24.42   & 21.13  & 26.04 & 26.40  & 22.26 & 31.47  & 28.96 & 30.79 & 31.69 & 27.02\\
    Ours-ALS   &24.50   & 21.17  & 25.99 & 26.46  & 22.26 & 31.52 & 28.90  & 30.73 & 31.68 & 27.02 \\
    \hline
    HR-3DGS    &24.41 &20.59 &26.58 &26.28 &22.27 &31.52 &29.12 &31.57 &32.36 &27.19 \\
    \hline
    \end{tabular}
\end{table*}

\begin{table*}[t!]
    \caption{Per-scene SSIM comparison on the Mip-NeRF 360 dataset ($\times8$ $\rightarrow $$\times2$). Ours-ALS refers to our method using adaptive-length subsequencing (ALS).}
    \label{tab:per-scene_lpips}
    \centering
    \begin{tabular}{|l|ccccc|cccc|c|}
    \hline
    & bicycle & flowers & garden & stump & treehill & room & counter & kitchen & bonsai & average \\
    \hline
    Bicubic         &0.6401 &0.5321 &0.6648 &0.7324 &0.5880 &0.8877 &0.8573 &0.8128 &0.8980 &0.7348  \\
    SwinIR + 3DGS   &0.6810 &0.5498 &0.7259 &0.7468 &0.6020 &0.9063 &0.8837 &0.8724 &0.9235 &0.7657  \\
    Ours-S          &0.6752 &0.5512 &0.7476 &0.7481 &0.6048 &0.9123 &0.8936 &0.9071 &0.9328 &0.7747 \\
    Ours-ALS        &0.6783 &0.5503 &0.7462 &0.7467 &0.6028 &0.9123 &0.8918 &0.9062 &0.9323 &0.7741  \\
    \hline
    HR-3DGS    &0.7007 &0.5445 &0.8173 &0.7571 &0.6269 &0.9263 &0.9144 &0.9325 &0.9465 &0.7962  \\
    \hline
    \end{tabular}
\end{table*}

\begin{table*}[t!]
    \caption{Per-scene LPIPS comparison on the Mip-NeRF 360 dataset ($\times8$ $\rightarrow $$\times2$). Ours-ALS refers to our method using adaptive-length subsequencing (ALS).}
    \label{tab:per-scene_lpips}
    \centering
    \begin{tabular}{|l|ccccc|cccc|c|}
    \hline
    & bicycle & flowers & garden & stump & treehill & room & counter & kitchen & bonsai & average \\
    \hline
    Bicubic                 &0.3688 &0.4315 &0.3469 &0.3334 &0.4391 &0.2750 &0.2671 &0.2598 &0.2392 &0.3290  \\
    SwinIR + 3DGS   &0.3220 &0.4065 &0.2784 &0.3098 &0.4116 &0.2354 &0.2216 &0.1973 &0.2035 &0.2873  \\
    Ours-S        &0.3344 & 0.4091 & 0.2613 & 0.3142 & 0.4162 & 0.2218 & 0.2074 & 0.1536 & 0.1927 & 0.2790 \\
    Ours-ALS        &0.3261 &0.4062 &0.2607 &0.3117 &0.4134 &0.2218 &0.2104 &0.1542 &0.1925 &0.2774 \\
    \hline
    HR-3DGS    &0.3230 &0.4188 &0.1777 &0.3130 &0.3997 &0.1931 &0.1800 &0.1136 &0.1758 &0.2550

  \\
    \hline
    \end{tabular}
\end{table*}

\begin{figure*}[h]
\centering
\includegraphics[width=\textwidth]{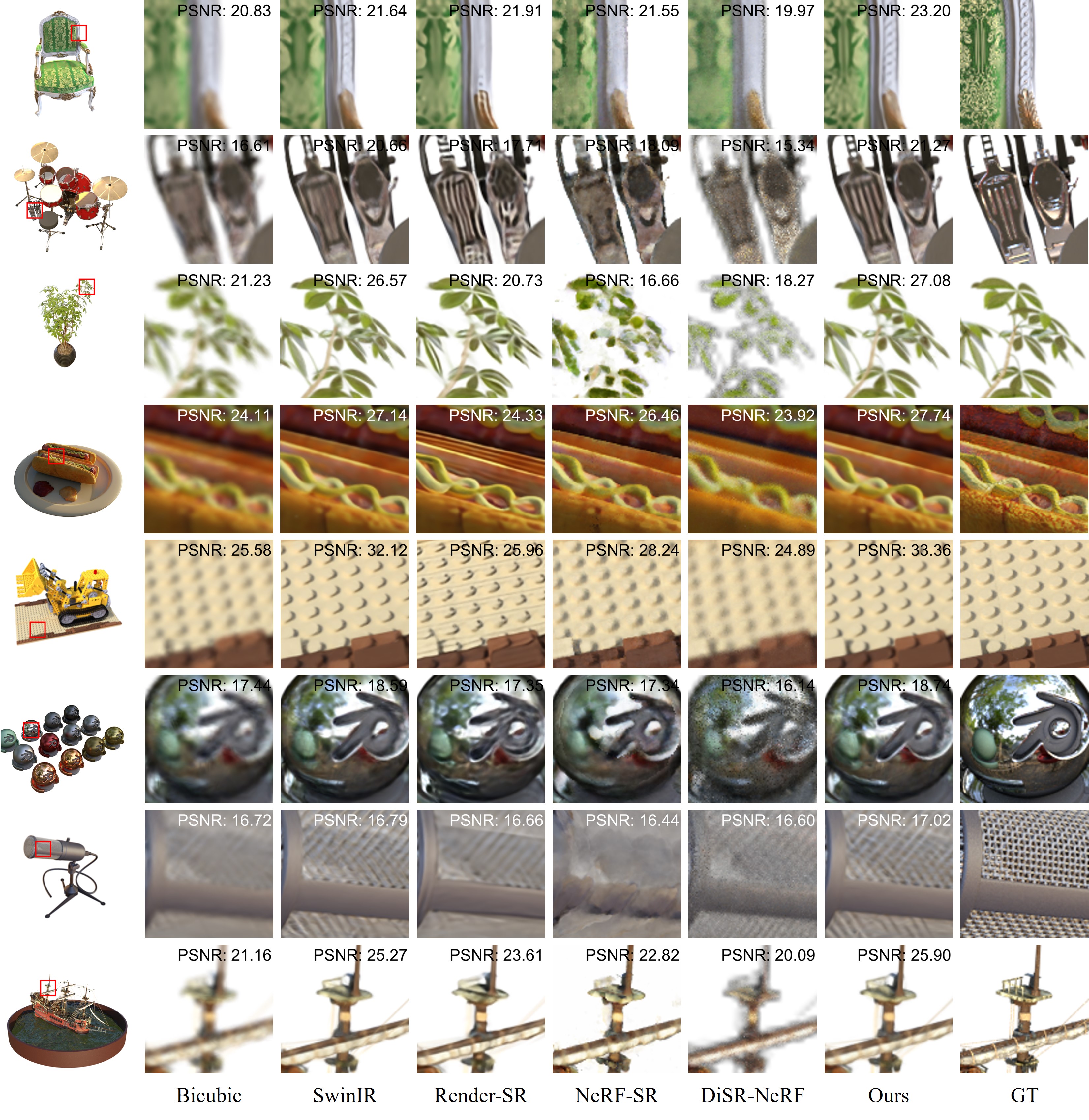} % Reduce the figure size so that it is slightly narrower than the column.

\caption{Qualitative results on the NeRF-synthetic dataset. The PSNR values against GT are embedded in each image patch. Ours have shown superior results than the existing baselines, especially for high-frequency details.}
\label{fig:appendix_qualitative_blender}
\end{figure*}

\begin{table}[ht]
   \centering
    \fontsize{9}{11}\selectfont
   \caption{Comparison with baseline models in Mip-NeRF 360 dataset ($\times8$ $\rightarrow$ $\times2$).}
   \label{tab:baseline_mip360}
   \begin{tabular}{|l|c|c|c|}
       \hline
       & PSNR$\uparrow$ & SSIM$\uparrow$ & LPIPS$\downarrow$\\
       \hline
       Bicubic                 & 26.22  & 0.7349 & 0.3290\\
       SwinIR     & 26.80 & 0.7657 & 0.2873\\
       SRGS$^{\dagger}$& 26.88  & 0.7670 & 0.2860 \\
        Ours                    & \textbf{27.02} & \textbf{0.7747}  & \textbf{0.2790}\\
       \hline
       3DGS-HR &27.19  &0.7710   &0.2802\\
       \hline
   \end{tabular}
\end{table}

\begin{table}[ht]
    \centering
    \caption{Temporal Consistency and Spatial Quality Metrics on Blender Dataset.}
    \label{tab:temporal_consistency}
    \begin{tabular}{|c||c|c|}
        \hline
        Method & FVD$\downarrow$ & PSNR$\uparrow$ \\
        \hline
        Bicubic & 195 & 27.56 \\
        SwinIR & 113 & 30.77 \\
        Render-SR & 134 & 28.90 \\
        NeRF-SR & 169 & 28.21 \\
        DiSR-NeRF & 304 & 26.00 \\
       \hline
        Ours-S & 110 & 31.35 \\
        Ours-ALS & \textbf{109} & \textbf{31.41} \\
        \hline
    \end{tabular}
\end{table}

\end{document}